\documentclass[journal, twocolumn]{IEEEtran}
\usepackage{enumerate}
\usepackage{amsmath}
\usepackage{booktabs}
\usepackage{multirow}
\usepackage{algorithm}
\usepackage{algorithmic}
\usepackage{subfigure}
\usepackage{amssymb}
\usepackage{mathrsfs}
\DeclareMathOperator*{\argmin}{arg\,min}

\usepackage{ifpdf}

\usepackage{cite}

\ifCLASSINFOpdf
  \usepackage[pdftex]{graphicx}
  \graphicspath{{../pdf/}{../jpeg/}}
  \DeclareGraphicsExtensions{.pdf,.jpeg,.png}
\else
  \usepackage[dvips]{graphicx}
  \graphicspath{{../eps/}}
  \DeclareGraphicsExtensions{.eps}
\fi
\usepackage{amsmath}
\usepackage{array}
\begin{document}
\bstctlcite{IEEEexample:BSTcontrol}
%
\title{Unsupervised Domain Adaptation with  Adversarial Residual Transform Networks}
%
%
%

\author{Guanyu Cai,
			Yuqin Wang,
			Lianghua He, and
			Mengchu Zhou, ~\IEEEmembership{Fellow, ~IEEE}
\thanks{Manuscript received July 16, 2018; revised December 11, 2018, and May 19, 2019; accepted August 2, 2019. This work was supported by the National Natural Science Foundation of China under Grant 61772369, 61773166, 61771144 and 61871004, Joint Funds of the National Science Foundation of China (U18092006), Shanghai Municipal Science and Technology Committee of Shanghai Outstanding Academic Leaders Plan (19XD1434000), Projects of International Cooperation of Shanghai Municipal Science and Technology Committee (19490712800). This paper is partially supported by the National Key R\&D Program 2018YFB1004701, by the Fundamental Research Funds for the Central Universities. (Corresponding author: Lianghua He). 
}
\thanks{G. Cai, Y. Wang and L. He are with the Department of Computer Science and Technology, Tongji University, Shanghai 201804, China (email: caiguanyu@tongji.edu.cn; wangyuqin@tongji.edu.cn; helianghua@tongji.edu.cn).}
\thanks{M. Zhou is with the Department of Electrical and Computer Engineering, New Jersey Institute of Technology, Newark, NJ 07102 USA, and also with the Institute of Systems Engineering, Macau University of Science and Technology, Macau 999078, China (e-mail: zhou@njit.edu).}
}

\maketitle

\begin{abstract}
Domain adaptation is widely used in learning problems lacking labels. Recent studies show that deep adversarial domain adaptation models can make markable improvements in performance, which include symmetric and asymmetric architectures. However, the former has poor generalization ability whereas the latter is very hard to train. In this paper, we propose a novel adversarial domain adaptation method named Adversarial Residual Transform Networks (ARTNs) to improve the generalization ability, which directly transforms the source features into the space of target features. In this model, residual connections are used to share features and adversarial loss is reconstructed, thus making the model more generalized and easier to train. Moreover, a special regularization term is added to the loss function to alleviate a vanishing gradient problem, which enables its training process stable. A series of experiments based on Amazon review dataset, digits datasets and Office-31 image datasets are conducted to show that the proposed ARTN can be comparable with the methods of the state-of-the-art.
\end{abstract}


%
\IEEEpeerreviewmaketitle

\section{Introduction}

\IEEEPARstart{D}{eep} neural networks trained on large-scale labeled datasets could achieve excellent performance across varieties of tasks, such as sentiment analysis~\cite{D14-1181,dos2014deep}, image classification~\cite{He_2016_CVPR,krizhevsky2012imagenet,ren2018deep} and semantic segmentation~\cite{long2015fully}. Yet they usually fail to generalize well on novel tasks because the transferability of features decreases as the distance between the base and target tasks increases~\cite{yosinski2014transferable}.  A convincing explanation is that there exists a domain shift between training data and testing one~\cite{ben2010theory,ben2007analysis}. To alleviate the negative effect caused by a domain shift, domain adaptation (DA) is proposed to utilize labeled data from a source domain to generalize models generalize well on a target domain~\cite{pan2010survey,shao2015transfer}.

Domain adaptation, which is a field belonging to transfer learning, has long been utilized to make it possible to exploit the knowledge learned in one specific domain to effectively improve the performance in a related but different domain. Earlier methods of DA aim to learn domain-invariant feature representations from data by jointly minimizing a distance metric that actually measures the adaptability between a pair of source and target domains, such as Transfer Component Analysis~\cite{pan2011domain}, Geodesic Flow Kernel~\cite{gong2012geodesic}, and Transfer Kernel Learning ~\cite{long2015domain}. In order to learn transferable features well, researchers apply deep neural networks to DA models~\cite{bengio2012deep,mesnil2011unsupervised,hinton2015distilling}. A feature extractor neural network is trained by reducing ``distance" between distributions of two different domains, on the assumption that the classifier trained by source data also works well in a target domain. In this kind of methods, Maximum Mean Discrepancy (MMD) loss is widely used for mapping different distributions~\cite{gretton2012optimal}. For example, Deep Adaptation Networks (DAN)~\cite{long2015learning}, Joint Adaptation Networks~\cite{pmlr-v70-long17a} and Residual Transfer Networks~\cite{Long:2016:UDA:3157096.3157112} apply MMD loss to several layers whereas Large Scale Detection through Adaptation~\cite{hoffman2014lsda} adds a domain adaptation layer that is updated based on MMD loss.

Recently, the idea of Generative Adversarial Networks (GANs)~\cite{goodfellow2014generative,wang2017generative} has been widely applied to DA. The methods of using GANs~\cite{liu2016coupled,bousmalis2017unsupervised} to transform source images to target ones are proposed and their classifiers are trained with the generated target images. However, when distributions of source and target domains are totally different, adversarial training has poor performance because of a gradient vanishing phenomenon. Alternative methods train GANs on features of  source and target domains. Their generator is acted as a feature extractor, and discriminator as a domain classifier. There are symmetric and asymmetirc adaptation architectures in adversarial domain adaptation, which can effectively adapt source and target distributions. The former's features in the source and target domains are generated from the same network~\cite{tzeng2015simultaneous,ganin2016domain}, while the latter's from different networks~\cite{tzeng2017adversarial}. It is well-recognized that the former is poor at generalization whereas the latter is difficult to train.

\begin{figure*}[htp]
\centering
\includegraphics[width = \textwidth]{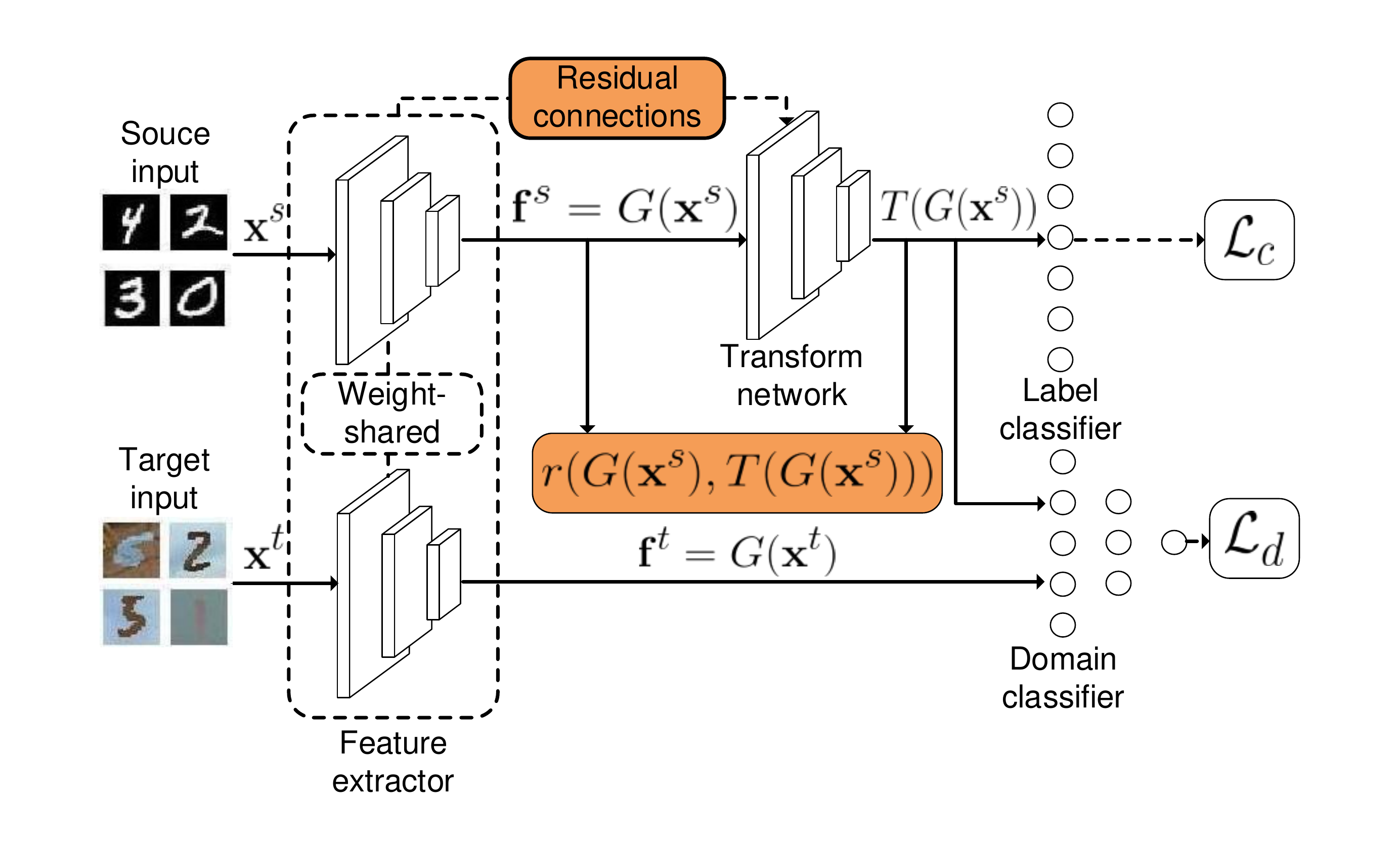}
\caption{The architecture of the proposed model. First feature extractor $G$ distills the feature representations of source and target samples. Then, transform network $T$ projects source features ${\bf f}^s=G({\bf x}^s)$ to the space of target features ${\bf f}^t=G({\bf x}^t)$. Finally, the label classifier $C$ is trained with the fake target features $T(G({\bf x}^s))$ and predicts the labels of target features $G({\bf x}^t)$ during a test period. In addition, domain classifier $D$ is trained to distinguish fake target features $T(G({\bf x}^s))$ and real target features $G({\bf x}^t)$, which can minimize the discrepancy between source and target domains through adversarial optimization. The regularization term $r(G({\bf x}^s),T(G({\bf x}^s)))$ measures the distance between ${\bf f}^s=G({\bf x}^s)$ and $T(G({\bf x}^s))$.}
\label{fig:flowchart}
\end{figure*} 

To solve the above problems, in this work, we propose a novel feature-shared model for adversarial domain adaptation, which achieves the flexibility of asymmetric architecture and can be easily trained. In the proposed framework as shown in Fig.~\ref{fig:flowchart}, a weight-shared feature extractor distills features from different domains, and a feature-shared transform network maps features from the source domain to the space of target features. Adversarial learning is completed with the losses from the label and domain classifiers. Note that we design residual connections between the extractor and the network to ease the learning of distribution mapping by sharing features. In addition, in order to avoid getting stuck into local minima, we construct a regularization term to ensure that the model at least knows a vague, if not exact, direction to match different distributions and overcome a gradient vanishing problem.
The main contributions of this work are as follows: 
\begin{enumerate}
\item A novel adversarial model that learns a non-linear mapping from a source domain to a target one is proposed. By using the features generated from the source domain, this model owns high generalization ability in the target domain.
\item During training, concise regularization that ensures the model to select the shortest path from all the transfer paths is constructed, thereby helping stabilize an adversarial optimization process.
\item This work extensively evaluates the proposed method on several standard benchmarks. The results demonstrate that our model outperforms the state-of-the-art methods in accuracy. Notably, our model can maintain excellent generalization and anti-noise abilities.
\end{enumerate}

Section~\ref{section2} reviews some related work on unsupervised domain adaptation. In Section~\ref{section3}, the proposed method is described. Several experiments are reported in Section~\ref{section4}. Section~\ref{section5} concludes this paper. 

\section{Related Work}
\label{section2}

DA has been extensively studied in recent years. Several studies give theoretical analyses of error bound when training and testing data are drawn from different distributions~\cite{ben2010theory,ben2007analysis}. This means that it is feasible to utilize the knowledge across domains. Moreover, the most important problem in DA is how to reduce the discrepancy between the source and target domains. Therefore, many methods modify a classifier to match different distributions~\cite{8334809,7999259,7723898,8337102}. However, the performances of these shallow methods are limited.  Recently, because deep neural networks can learn feature representations including information identification, and these features are also transferable~\cite{yosinski2014transferable}, they are widely used in DA methods. Most of them focus on how to measure the distance between different domains and how to design a network structure, and achieve remarkable performance.

\subsection{Traditional Domain Adaptation}

A general idea is to reweight instances in the source domain, where instances similar to the target distribution  are considered with more importance. This kind of method has proven effective in adaptation for the differences between the source and target distributions. In detail, the calculated weight is taken as the loss coffiecient of each source instance. Some methods reweight instances with direct importance estimation algorithms, such as~\cite{kanamori2009least,huang2007correcting,sugiyama2008direct,li2017prediction} and~\cite{7723898}. Other methods reweight instances with noisy labels to reduce the marginal and conditional shifts~\cite{zhang2013domain,liu2016classification}.

Another DA idea is to explicitly make source and target distributions similar. Many statistic characteristics are chosen to be the metrics to align subspaces of different distributions. MMD, which measures expectation difference  in reproducing kernel Hilbert space (RKHS) of source and target distributions, is widely used in many methods~\cite{pan2011domain,long2015domain,8337102,8334809}. In order to align more complex statistic characteristics, the studies~\cite{fernando2013unsupervised,sun2015subspace,sun2016return} calculate statistic moments of different order to match different distributions, which are easy to implement with low computational complexity. Instead of exploiting statistic characteristics, some methods~\cite{gong2012geodesic,cheng2014semi,courty2017optimal,courty2017joint} utilize manifold learning methods to transform a source distribution to a target one. In these methods, feature spaces are refined into low-dimensional spaces and feature distortion is thus avoided.

\subsection{Deep Domain Adaptation}

Recent, development of deep neural networks promotes deep domain adaptation. In~\cite{donahue2014decaf}, the experimental results demonstrate that features of deep neural networks instead of hand-crafted ones alleviate the negative influences of a domain shift even without any adaptations. However, a main limitation of using pre-trained deep features is that they severaly restrict the range of application. Later, a number of methods combine statistic characteristics with deep neural networks to a unified framework, which greatly improve performance on different tasks. In~\cite{long2015learning,pmlr-v70-long17a,Long:2016:UDA:3157096.3157112,
hoffman2014lsda}, MMD is embedded in deep convolutional neural networks. In~\cite{sun2016return} and~\cite{DBLP:journals/corr/ZellingerGLNS17}, high order moments are utilized to align feature spaces of source and target domains. 

Instead of designing fancy regularizers, some methods design special architectures to minimize the discrepancy between source and target domains. In~\cite{chopra2005learning}, a Siamese architecture is introduced to adapt pairs of source and target instances. In~\cite{glorot2011domain,kan2015bi}, auto-encoders are suggested to learn the transformation from a source domain to a target one.

Some methods choose adversarial loss to learn manifest invariant factors underlying different populations with a domain discriminator subnetwork. In these models, deep features are learned to confuse a domain discriminator such that they could capture the most discriminative information about classification instead of characteristics of domains. Domain-Adversarial Neural Networks (DANN)~\cite{ganin2016domain} consist of a symmetric feature generator, label discriminator, and domain discriminator. The whole model can be directly optimized via a gradient reversal algorithm. Deep Reconstruction-Classification Networks~\cite{ghifary2016deep} also take adversarial learning and add a reconstruction step for target images. Adversarial Discriminative Domain Adaptation (ADDA)~\cite{tzeng2017adversarial} uses an asymmetric feature generator that is trained alternatively with a domain classifier. 

The above-mentioned domain adversarial networks fall into two categories. Some methods, such as domain confusion networks~\cite{tzeng2015simultaneous} and DAN~\cite{ganin2016domain}, share weights between source and target feature extractors. They use the same network to learn representations from different inputs, which learns a symmetric transformation to utilize the transferability of features generated from deep neural networks and reduces parameters in the model. Other methods construct two networks for source and target domains, respectively~\cite{tzeng2017adversarial,liu2016coupled,bousmalis2017unsupervised}. They can learn an asymmetric transformation, allowing networks to learn parameters for each domain individually. In theory, asymmetric transformation can lead to more effective adaptation~\cite{rozantsev2018beyond}.

Adversarial domain adaptation has also been explored in generative adversarial networks (GANs). Coupled Generative Adversarial Networks (CoGANs)~\cite{liu2016coupled} apply GANs to DA. Two GANs are trained to generate source and target images, respectively. Pixel-Level Domain Adaptation~\cite{bousmalis2017unsupervised} uses a conditional GAN model to synthesize target images to facilitate training a label classifier.  Methods based on GANs can improve the performance of digits datasets, but their downside is a difficult training process as caused by gradient vanishing when facing more natural image datasets according to~\cite{arjovsky2017wasserstein}. In this work, we focus on learning the mapping of different feature spaces instead of synthesizing target images, and propose a discriminative model aiming to adapt distinct domains.

\section{Adversarial Residual Transform Networks}
\label{section3}

In this section, we describe the details of our proposed
model. We first define unsupervised domain adaptation and preliminary domain adversarial networks, and then demonstrate the key innovations of our model, which can well handle the problems encountered by the previous models. At last, we give a complete algorithm of matching the distributions of target and source domains using our model.

\subsection{Definitions}

When it comes to a machine learning task, a domain $D$ corresponds to four parts: feature space $\mathcal{X}$, label space $\mathcal{Y}$, marginal probability distribution $P({\bf X})$ and conditional probability distribution $P({\bf X}|{\bf Y})$, where ${\bf X}\in \mathcal{X}$, ${\bf Y}\in\mathcal{Y}$. Subscript $s$ and $t$ are used to denote the source and target distributions. In a traditional machine learning task,  training data are drawn from source domain $D_s$ and testing data are drawn from target domain $D_t$, where their marginal and conditional probability distributions are the same ($P_s({\bf X}^s)=P_t({\bf X}^t), P_s({\bf X}^s|{\bf Y}^s)=P_t({\bf X}^t|{\bf Y}^t)$). Thus, models trained in the source domain are feasible to the target one. However, in unsupervised DA, these assumptions are not valid, which leads us to a more difficult problem as follows.

Given a source domain as $D_s\{{\bf x}^s_i,{\bf y}^s_i\}_{i=1}^{n_s}$, where $n_s$ is the number of source domain samples, ${\bf x}^s_i$ is the $i$th instance in $D_s$ and ${\bf y}^s_i$ is the label of ${\bf x}^s_i$. Similarly, a target domain is denoted as $D_t\{{\bf x}^t_i\}_{i=1}^{n_t}$, where $n_t$ is the number of target domain samples, ${\bf x}^t_i$ is the $i$th instance in $D_t$. The source and target domains are drawn from distribution $P_s({\bf X}^s)$ and $P_t({\bf X}^t)$, respectively, which are different. In most cases, conditional probability distributions are also different ($P_s({\bf X}^s|{\bf Y}^s)\neq P_t({\bf X}^t|{\bf Y}^t)$) The goal is to learn a feature extractor $G_t$ and a classifier $C_t$ for $D_t$. $G_t$ distills feature representations ${{\bf f}^t=G_t({\bf x}^t)}$ from target samples, and $C_t$ correctly predicts the labels of target samples receiving ${{\bf f}^t=G_t({\bf x}^t)}$. Because of lacking annotations in $D_t$, DA learns $G_s$ and $C_s$ with samples from $D_s$, and tries to adapt them to be useful in $D_t$.

\subsection{Adversarial Domain Adaptation}

To solve an unsupervised DA problem, a number of methods have been proposed. Among the most effective ones is  adversarial domain adaptation. This work aims to modify this kind of framework to improve its generalization and anti-noise ability. In DA problems, it is difficult to train $G_t$ and $C_t$ for a target domain without labels. However, because there exists a correlation between the source and target domains, it is common to utilize $G_s$ and $C_s$ to predict labels of target samples. In order to make $G_s$ and $C_s$ valid in a target domain, adversarial DA models are usually used to train a feature extractor $G$, a label classifier $C$ and a domain classifier $D$ for both domains. In details, these models set $G=G_t=G_s$, and $C=C_t=C_s$, which means the feature extractor and label classifier are used for both source and target domains. Specifically, $D$ also receives feature representations from $G$ and is trained to minimize the discrepancy between source and target feature distributions: $G({\bf x}^s)$ and $G({\bf x}^t)$. An adversarial training procedure is a minimax two-player game~\cite{goodfellow2014generative}. One player $D$ learns to distinguish whether features are from a source or target domain, whereas the other $G$ tries to generate domain-invariant features. They have contradictory optimization objectives, and their objectives are optimized alternately in this minmax game. To train the whole network in an end-to-end way, DANN~\cite{ganin2016domain} adopts the following loss function:
\setlength{\arraycolsep}{0.0em}
\begin{align}
{\cal L}(\theta_d,\theta_g,\theta_c)=&\frac{1}{n_s}\sum_{{\bf x}_i\in D_s}{\cal L}_c(C(G({\bf x}_i)),y_i)- \nonumber \\
&\frac{\lambda}{n}\sum_{{\bf x}_i\in D_s\cup D_t}{\cal L}_d(D(G({\bf x}_i)),d_i)\label{eq1}
\end{align}
where $n=n_s+n_t$, and ${\cal L}_c$ and ${\cal L}_d$ denote losses of $C$ and $D$, respectively. $\lambda$ is a trade-off parameter between ${\cal L}_c$ and ${\cal L}_d$. $\theta_d$, $\theta_g$ and $\theta_c$ are the parameters of $D$, $G$ and $C$, respectively. $y_i$ and $d_i$ denote the class and domain labels of images. After convergence, optimal parameters $\hat{\theta}_d$, $\hat{\theta}_c$ and $\hat{\theta}_g$ can deliver a saddle point given as:
\setlength{\arraycolsep}{0.0em}
\begin{align}
\hat{\theta}_d=& \argmin_{\theta_d}{\cal L}_d(\theta_d,\theta_g)\label{eq2}\\
\hat{\theta}_c=& \argmin_{\theta_c}{\cal L}_c(\theta_g,\theta_c)\label{eq3}\\
\hat{\theta}_g=& \argmin_{\theta_g}{\cal L}(\theta_d, \theta_g,\theta_c)\label{eq4}
\end{align}
\setlength{\arraycolsep}{5pt}

In such framework, a DA model can be trained in an end-to-end way.  
The intuitive idea behind this model is that with the minmax two-player game going, $D$ and $G$ strengthen each other. When the training procedure converges, the features of different domains generated from $G$ are very hard to distinguish by $D$. In this condition, features are domain-invariant and the feature distributions of different domains are adapted.

Theoretically, adversarial domain adaptation is based on $\mathcal{H}$-divergence in~\cite{ben2007analysis,ben2010theory}.  
However, it is almost impossible to apply $\mathcal{H}$-divergence in real-world algorithms. Because it is defined in a binary classification problem and requires a global search in all hypothesis. In~\cite{ben2007analysis}, an approximate algorithm is given. Given a generalization error $\epsilon$ of discriminating between source and target instances, $\mathcal{H}$-divergence is computed as:
\begin{equation}
\hat{d}_{\mathcal{A}}=2(1-2\epsilon)
\end{equation}

The value of $\hat{d}_{\mathcal{A}}$ is called the {\em Proxy $\mathcal{A}$-distance} (PAD).  
In adversarial domain adaptation, the domain classifier composed of neural networks is trained to directly decrease PAD. Cooperating with the domain classifier, the feature extractor learns domain-invariant features from different domains, implying that discrepancy between source and target distributions is decreased. Several models based on this kind of architecture have achieved the top performances in different visual tasks~\cite{ganin2016domain,ghifary2016deep}.

\subsection{Residual Connections}

The proposed method does not rely on only feature extractor $G$ to map different distributions. Instead, we construct an adversarial residual transform network (ARTN) $T$ to project source features ${\bf f}^s=G({\bf x}^s)$ to the space of target features. The network is trained to generate fake target features $T(G({\bf x}^s))$, which are in the same distribution as real target features ${\bf f}^t=G({\bf x}^t)$. Then, we use the fake target features $T(G({\bf x}^s))$ and corresponding labels ${\bf y}^s$ to train a classifier $C$ for the target domain. After training, the labels of target samples are predicted by $C$. 

In previous unsupervised DA methods, the weights of feature extractor $G$ for source and target domains are shared~\cite{long2015learning,pmlr-v70-long17a,Long:2016:UDA:3157096.3157112,
hoffman2014lsda}.  
However, regarding matching different distributions, the generalization ability of asymmetric transformation is better than that of symmetric one~\cite{tzeng2017adversarial}.  
If the networks are trained to capture domain-invariant information from source features and utilize them to classify target samples, there would be a boost to their generalization ability. 
However, the asymmetric architecture proposed in~\cite{tzeng2017adversarial} is hard to obtain such enhancement and the feature extractor for a target domain is easy to collapse, because there exists no relationship between the feature extractors of source and target domains. In order to make our model learn domain-invariant information and avoid diverging during its training, we propose a transform network that builds connections between source and target domain features. 

\begin{figure}
  \centering
  \subfigure[Source samples]{
    \label{fig:subfig:residual1} 
    \includegraphics[width=0.85\columnwidth]{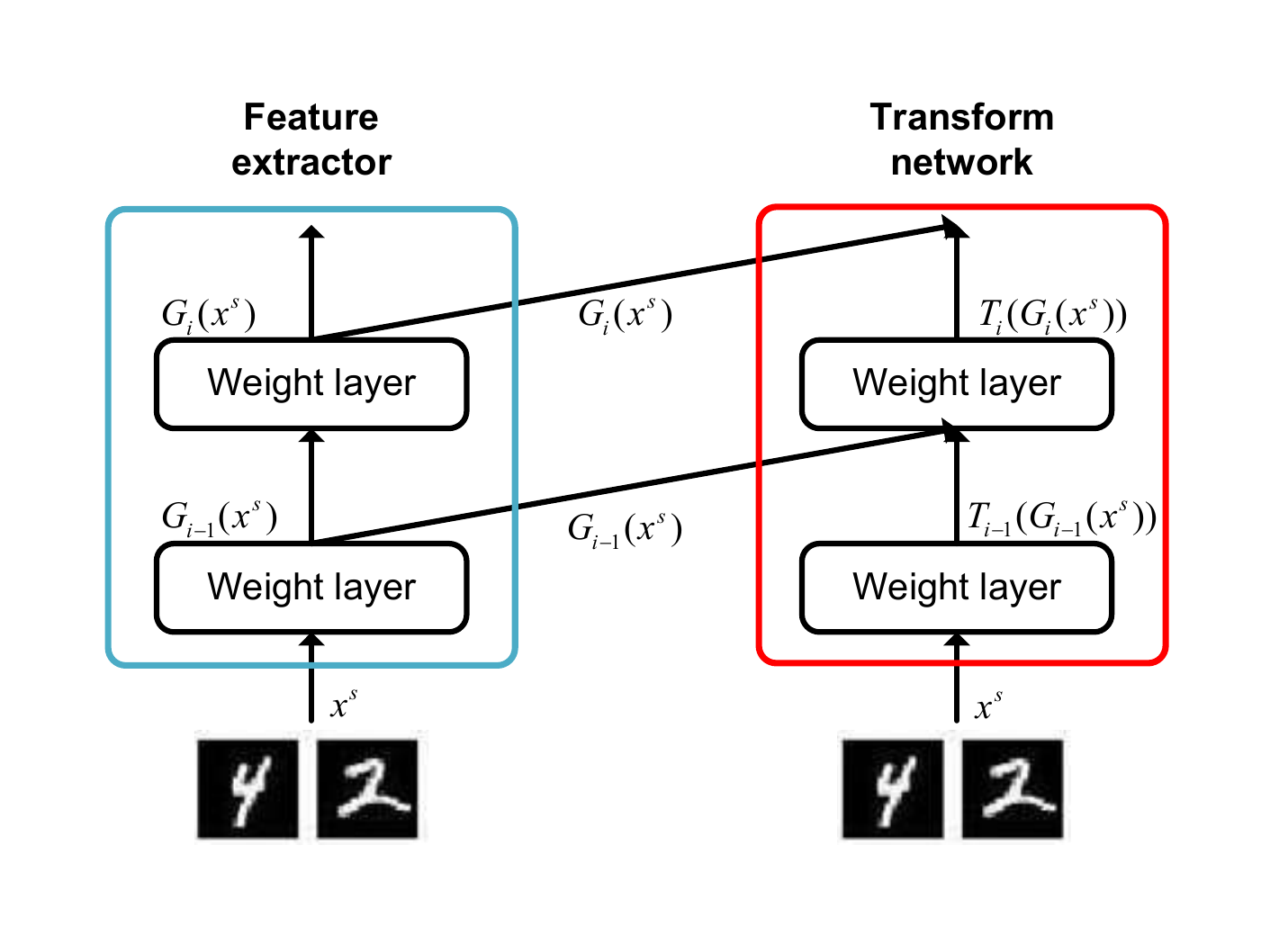}}
  \hspace{1in}
  \subfigure[Target samples]{
    \label{fig:subfig:residual2} 
    \includegraphics[width=0.85\columnwidth]{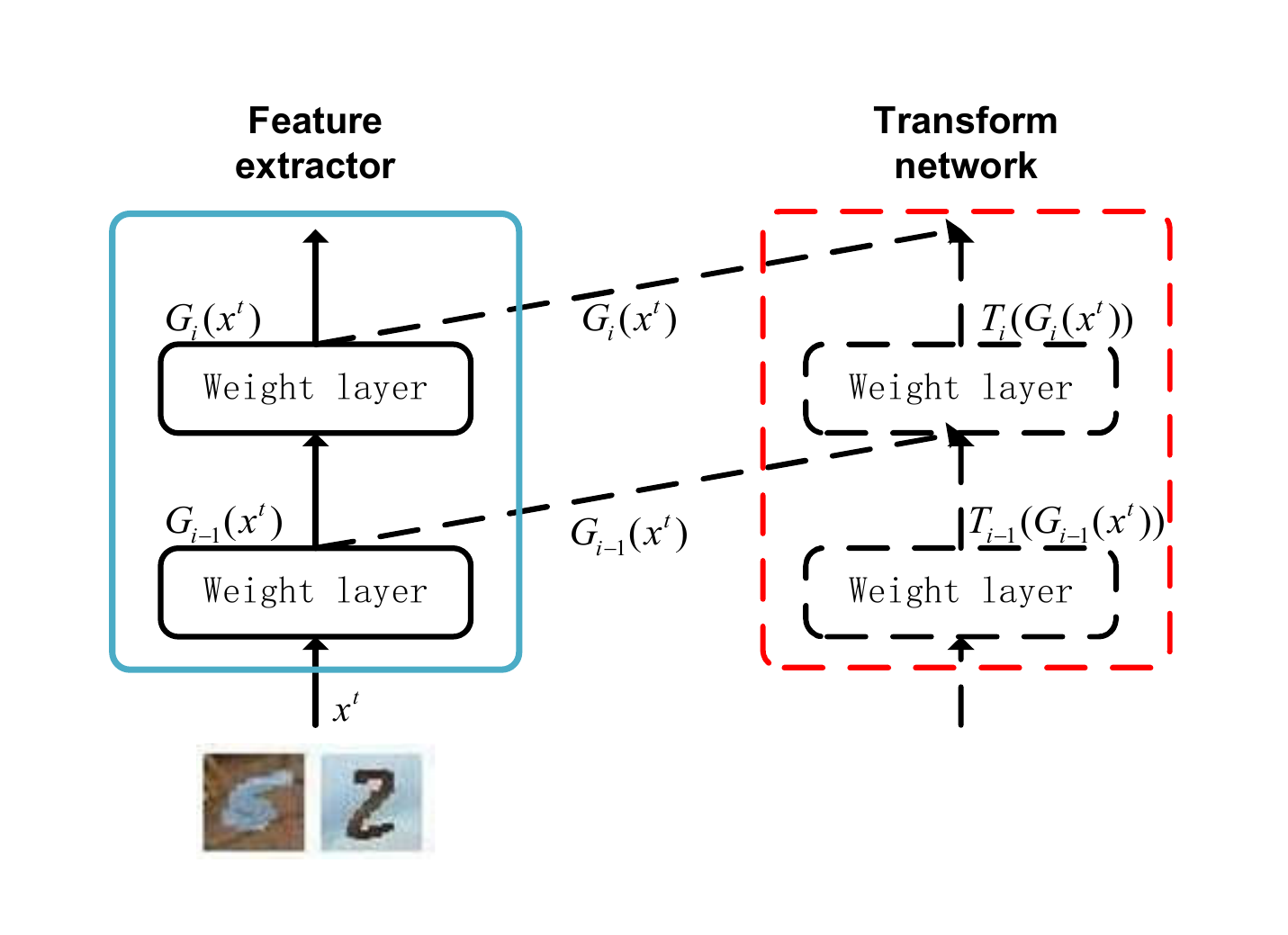}}
\caption{Residual connections between the feature extractor and transform network. When the inputs are source samples, the feature extractor and transform network are both activated (the solid line represents that this network is activated, whereas the dashed line is opposite) and the features distilled from a feature extractor would be conveyed to a transform network. When the inputs are target samples, only the feature extractor is activated. The features are not shared between the feature extractor and transform network.}
  \label{fig:residual} 
\end{figure}

The detailed architecture of residual connections between a feature extractor and a transform network is shown in Fig.~\ref{fig:residual}. The weight-tied feature extractor $G$ is trained to capture representations from source samples ${\bf x}^s$ and target samples ${\bf x}^t$. The transform network stacks a few layers by using the same architecture with a feature extractor. Unlike the symmetric transformation, the proposed network shares features with the feature extractor instead of parameters. Our network is also different from asymmetric transformation where two networks have no relationship. We add residual connections between the feature extractor and the transform network to share features. Therefore, with a carefully designed architecture, our model is able to alleviate the drawbacks of both symmetric and asymmetric models, which is never seen in the literature to our best knowledge.

Theoretically, by denoting a desired underlying mapping between source and target distributions as $M$ and letting $G({\bf x}^t)=M(G({\bf x}^s))$, we intend to train transform network $T$ to fit a mapping of $T(G({\bf x}^s))=M(G({\bf x}^s))-G({\bf x}^s)$. In details, for an $N$ layer transform network, the $i$th layer of a transform network where $i<N$ is defined as:
\begin{equation}
T_i(G_i({\bf x}^s))=T_{i}(T_{i-1}(G_{i-1}({\bf x}^s)))+G_{i}({\bf x}^s)
\end{equation}
where $T_i(\cdot)$ and $G_i(\cdot)$ denote the $i$th layer of the transform network and feature extractor individually. The inputs of $T$ are original data, and the output is $T_N(G_N({\bf x}^s))$. 

Please note that residual connections in different methods tend to realize different purposes.  
Firstly, the effect of residual connections in our paper is different from others. For example, residual connections in ResNet~\cite{He_2016_CVPR} are used to shorten their training process by making gradients flow well. However, residual connections in U-Net~\cite{ronneberger2015u} are used to enable precise localization for segmentation. Although these papers all utilize skip connections, they still make novel contributions. In our paper, residual connections are used to capture semantic information from a source domain. This modification has not been seen in other domain adaptation methods and the effect is also different from papers in other research areas. Secondly, the detailed modification is different from skip connections in other papers. Skip connections in ResNet and U-Net are constructed in a single network across different layers whereas ours are constructed in the same layer across different networks.

\subsection{Vanishing Gradient Problem in Adversarial Training}

The detailed theoretical derivation and training process of adversarial DA has been described in~\cite{ganin2016domain}. Yet there exists a vanishing gradient problem in adversarial training. In this section, its theoretical analysis is presented.

Once we adopt a transform network in adversarial DA and utilize cross entropy loss function for $D$, the adversarial nets-based DA framework of $D$ and $G$ needs the following minmax optimization:
\begin{align}
\min_{G,T}\max_{D}{\cal L}(D,G,T)=&{\mathbb{E}}_{{\bf x}\thicksim P_s}[logD(T(G({\bf x})))]+\nonumber\\
&{\mathbb{E}}_{{\bf x}\thicksim P_t}[log(1-D(G({\bf x})))]\label{eqd-1}
\end{align}
where maximizing the loss with respect to $D$ yields a tighter lower bound on the true domain distribution divergence, whereas minimizing the loss with respect to $G$ and $T$ minimizes the distribution divergence in the feature space.

For any given $G$ and $T$, the optimal $D^{*}$ is obtained at:
\begin{equation}
D^{*}({\bf z})=\frac{P_s({\bf z})}{P_s({\bf z})+P_t({\bf z})}
\end{equation}
where ${\bf z}$ is the sample in the feature space. For ${\bf z}\thicksim P_s$, ${\bf z}=T(G(\bf x))$, while for ${\bf z}\thicksim P_t$, ${\bf z}=G({\bf x})$. Similar to~\cite{goodfellow2014generative}, we give the proof as follows.

\emph{Proof.} For any given $G$ and $T$, the training criterion for $D$ is to maximize ${\cal L}(D,G,T)$:
\begin{align}
\max_{D}{\cal L}(D,G,T)&=\int_{\bf x}P_s({\bf x})logD(T(G({\bf x})))+\nonumber\\
&P_t({\bf x})log(1-D(G({\bf x})))d{\bf x}\nonumber\\
&=\int_{\bf z}P_s({\bf z})logD({\bf z})+\nonumber\\
&P_t({\bf z})log(1-D({\bf z}))d{\bf z}
\end{align}
We take the partial differential of ${\cal L}(D,G,T)$ with respect to $D$, and achieve its maximum in $[0,1]$ at $D^{*}({\bf z})=\frac{P_s({\bf z})}{P_s({\bf z})+P_t({\bf z})}$.

Given $D^{*}$, the minmax optimization can now be reformulated as:
\begin{align}
\min_{G,T}{\cal L}(D^*,G,T)&={\mathbb{E}}_{{\bf z}\thicksim P_s}[logD^*({\bf z})]+\nonumber\\
&{\mathbb{E}}_{{\bf z}\thicksim P_t}[log(1-D^*({\bf z}))]\nonumber\\
&={\mathbb{E}}_{{\bf z}\thicksim P_s}[log\frac{P_s({\bf z})}{P_s({\bf z})+P_t({\bf z})}]+\nonumber\\
&{\mathbb{E}}_{{\bf z}\thicksim P_t}[log\frac{P_t({\bf z})}{P_s({\bf z})+P_t({\bf z})}]\nonumber\\
&={\mathbb{E}}_{{\bf z}\thicksim P_s}[log\frac{2P_s({\bf z})}{P_s({\bf z})+P_t({\bf z})}]+\nonumber\\
&{\mathbb{E}}_{{\bf z}\thicksim P_t}[log\frac{2P_t({\bf z})}{P_s({\bf z})+P_t({\bf z})}]-2log2\nonumber\\
&=2\cdot JSD(P_s||P_t)-2log2\label{eq12}
\end{align}
where $JSD(\cdot)$ is the Jensen-Shannon divergence. Since the Jensne-Shannon divergence between two distributions is always non-negative, and zero if they are equal, ${\cal L^*}=-2log2$ is the global minimum of  ${\cal L}(D,G,T)$ where the only solution is $P_s=P_t$. In this case, the distributions of source and target domains are the same and the goal of DA is well achieved.

However, in practice, adversarial DA remains remarkably difficult to train. It is sensitive to the initializatoin of parameters and its training process tend to be unstable, i.e., ${\cal L}(D,G,T)$ does not converge. These problems are caused by a vanishing gradient phenomenon. In theory, Jensen-Shannon divergence measures the difference between source and target distritbutions are different. By minimizing it, source and target distributions in the feature space tend to be the same. However, if we utilize a gradient descent algorithm to  optimize ${\cal L}(D,G,T)$ which is the most common algorithm for nerual networks, Jensen-Shannon divergence is difficult to converge because its gradient is easily stuck into zero, to be proved next.

According to~\cite{arjovsky2017wasserstein}, $P_s$ and $P_t$ can be regarded as two distributions that have support contained in two closed manifolds ${\mathcal M}$ and ${\mathcal N}$ that do not have full dimension, respectively. $P_s$ and $P_t$ are continuous in their respective manifolds, which means that if a set $A$ with measure 0 in ${\mathcal M}$, then $P_s(A)=0$. In this case, $JSD(P_s||P_t)=log2$ for almost any $P_s$ and $P_t$. We need to use Lemma 3.1~\cite{arjovsky2017wasserstein} to prove it:

\newtheorem{lemma}{Lemma}[section]
\begin{lemma} \label{lemma1}
\emph {Let ${\mathcal M}$ and ${\mathcal P}$ be two regular submanifolds that do not perfectly align and do not have full dimension. Let ${\mathcal L}={\mathcal M}\cap{\mathcal P}$. If ${\mathcal M}$ and ${\mathcal P}$ do not have boundary, then ${\mathcal L}$ is also a manifold. and has strictly lower dimension than both of ${\mathcal M}$ and ${\mathcal P}$. If they have boundary,
${\mathcal L}$ is a union of at most 4 strictly lower dimensional manifolds. In both cases, ${\mathcal L}$ has measure 0 in
both ${\mathcal M}$ and ${\mathcal P}$}
\end{lemma} 

\emph {Proof.} By Lemma 3.1, we know that ${\mathcal L}={\mathcal M}\cap{\mathcal P}$ has strictly lower dimensional than both ${\mathcal M}$ and ${\mathcal P}$ do, such that $P_s({\mathcal L})=0$ and $P_t({\mathcal L})=0$.
\begin{align}
2\cdot JSD(P_s||P_t)&=\int_{\bf z}P_s({\bf z})log\frac{2P_s({\bf z})}{P_s({\bf z})+P_t({\bf z})}+\nonumber\\&P_t({\bf z})log\frac{2P_t({\bf z})}{P_s({\bf z})+P_t({\bf z})}d{\bf z}\nonumber\\
&=\int_{\bf z\in {\mathcal M}\setminus{\mathcal L}}P_s({\bf z})log\frac{2P_s({\bf z})}{P_s({\bf z})+P_t({\bf z})}+\nonumber\\&P_t({\bf z})log\frac{2P_t({\bf z})}{P_s({\bf z})+P_t({\bf z})}d{\bf z}\nonumber\\
&+\int_{\bf z\in {\mathcal N}\setminus{\mathcal L}}P_s({\bf z})log\frac{2P_s({\bf z})}{P_s({\bf z})+P_t({\bf z})}+\nonumber\\&P_t({\bf z})log\frac{2P_t({\bf z})}{P_s({\bf z})+P_t({\bf z})}d{\bf z}\nonumber\\
&+\int_{\bf z\in {\mathcal L}}P_s({\bf z})log\frac{2P_s({\bf z})}{P_s({\bf z})+P_t({\bf z})}+\nonumber\\&P_t({\bf z})log\frac{2P_t({\bf z})}{P_s({\bf z})+P_t({\bf z})}d{\bf z}\nonumber\\
&+\int_{\bf z\in \left({\mathcal M}\cup{\mathcal N}\right)^c}P_s({\bf z})log\frac{2P_s({\bf z})}{P_s({\bf z})+P_t({\bf z})}+\nonumber\\&P_t({\bf z})log\frac{2P_t({\bf z})}{P_s({\bf z})+P_t({\bf z})}d{\bf z}
\end{align}
where $\left({\mathcal M}\cup{\mathcal N}\right)^c$ is the complement of $\left({\mathcal M}\cup{\mathcal N}\right)$. For ${\bf z\in {\mathcal M}\setminus{\mathcal L}}$, $P_s({\bf z})=1$ and $P_t({\bf z})=0$. Similarly, for ${\bf z\in {\mathcal N}\setminus{\mathcal L}}$, $P_t({\bf z})=1$ and $P_s({\bf z})=0$. When ${\bf z\in ({\mathcal M}\cup{\mathcal L})^c}$ and ${\bf z\in {\mathcal L}}$, $P_s({\bf z})$ and $P_t({\bf z})$ are equal to zero. Therefore, $JSD(P_s||P_t)=log2$.

Note that when $JSD(P_s||P_t)$ is a constant, gradients for all the parameters in an adversarial DA network are zeros. Therefore, if a gradient descent algorithm is adopted, the vanishing gradient problem appears and divergence between source and target domains is difficult and sometimes impossible to be minimized.

\subsection{Regularizer Based on Transport Theory}

Once we have parametrized $G$ and $T$, we employ adversarial loss to adapt different distributions. The architecture modification requires us to revise our loss function. Instead of measuring the distance between source features ${\bf f}^s=G({\bf x}^s)$ and target features ${\bf f}^t=G({\bf x}^t)$ generated from one feature extractor, the proposed model lets domain classifier $D$ discriminate source features ${\bf f}^s=T(G({\bf x}^s))$ from the transform network and target features ${\bf f}^t=G({\bf x}^t)$ from the feature extractor. Thus, the loss function is modified from (\ref{eq1}) into:
\begin{align}
{\cal L}(\theta_d,\theta_g,\theta_c, \theta_t)=&\frac{1}{n_s}\sum_{{\bf x}_i\in D_s}{\cal L}_c(C(T(G({\bf x}_i))),y_i)- \nonumber \\
&\frac{\lambda}{n}(\sum_{{\bf x}_i\in D_s}{\cal L}_s(D(T(G({\bf x}_i))),d_i^s)+ \nonumber \\
&\sum_{{\bf x}_i\in D_t}{\cal L}_t(D(G({\bf x}_i))),d_i^t))\label{eq6}
\end{align}
where $d_i^s$ and $d_i^t$ denote the domain labels of the $i$th source and target samples, respectively. ${\cal L}_s$ and ${\cal L}_t$ denote the domain loss of source and target samples, respectively. $\theta_t$ denotes the parameters of $T$. This objective funtion replaces $G({\bf x}_i)$ in (\ref{eq1}) with $T(G({\bf x}_i))$, indicating that our model uses features generated from transform network $T$ to be the input of label classifier $C$ and domain classifier $D$.

As our proof, if we optimize ${\cal L}(\theta_d,\theta_g,\theta_c, \theta_t)$ as general adversarial DA framework, a vanishing gradient problem would disturb us. To address this issue, we add a regularization term to the loss function based on the optimal transport problem as defined by Monge~\cite{courty2017optimal}. DA's goal is to find a mapping from a source domain to a target one, while the optimal transport problem gives a solution that transfers one distribution to another. Therefore, that problem can be represented in the form of Monge's formulation of the optimal transport problem~\cite{courty2017optimal,courty2017joint}. If we denote the probability measures over $P_s$ and $P_t$ as $\mu_s$ and $\mu_t$, respectively, Monge's formulation of the optimal transport problem is:
\begin{equation}
\label{eq7}
T_0=\argmin_T\int_{{\bf x}\in P_s}r({\bf x},T({\bf x}))d\mu({\bf x}), s.t. T\#(\mu_s)=\mu_t
\end{equation}
where $r(\cdot)$ denotes some kind of distance metric, $T$ denotes a transport mapping from $P_s$ to $P_t$, and $T_0$ is the optimal solution of $T$. $T\#(\mu_s)$ denotes the push forward of $\mu_s$ by a measureable function $T$. ${\bf x}$ denotes the samples drawn from $P_s$. DA's goal is to find a transport mapping $T_0$ satifying $T\#(\mu_s)=\mu_t$, which means that a transformation from source distribution $P_s$ to target distribution $P_t$ should be found. Specifically, in our model, we use transform network $T$ to fit the transport mapping to meet $T\#(\mu_s)=\mu_t$ via adversarial training. By fitting $r(\cdot)$, we can construct a regularization term that measures the distance between $G({\bf x}^s)$ and $T(G({\bf x}^s))$. In our model, according to our empirical evaluation results, $r(\cdot)$ is the cosine distance between them:
\begin{equation}
\label{eq8}
r(G({\bf x}^s),T(G({\bf x}^s)))=-\frac{\langle G({\bf x}^s)\cdot T(G({\bf x}^s))\rangle}{\lvert G({\bf x}^s)\rvert \cdot \lvert T(G({\bf x}^s))\rvert}
\end{equation}
where $\langle \cdot \rangle$ denotes an inner product, and $\lvert \cdot \rvert$ denotes $L_2$ norm.

\begin{figure}[htp]
\centering
\includegraphics[width = \columnwidth]{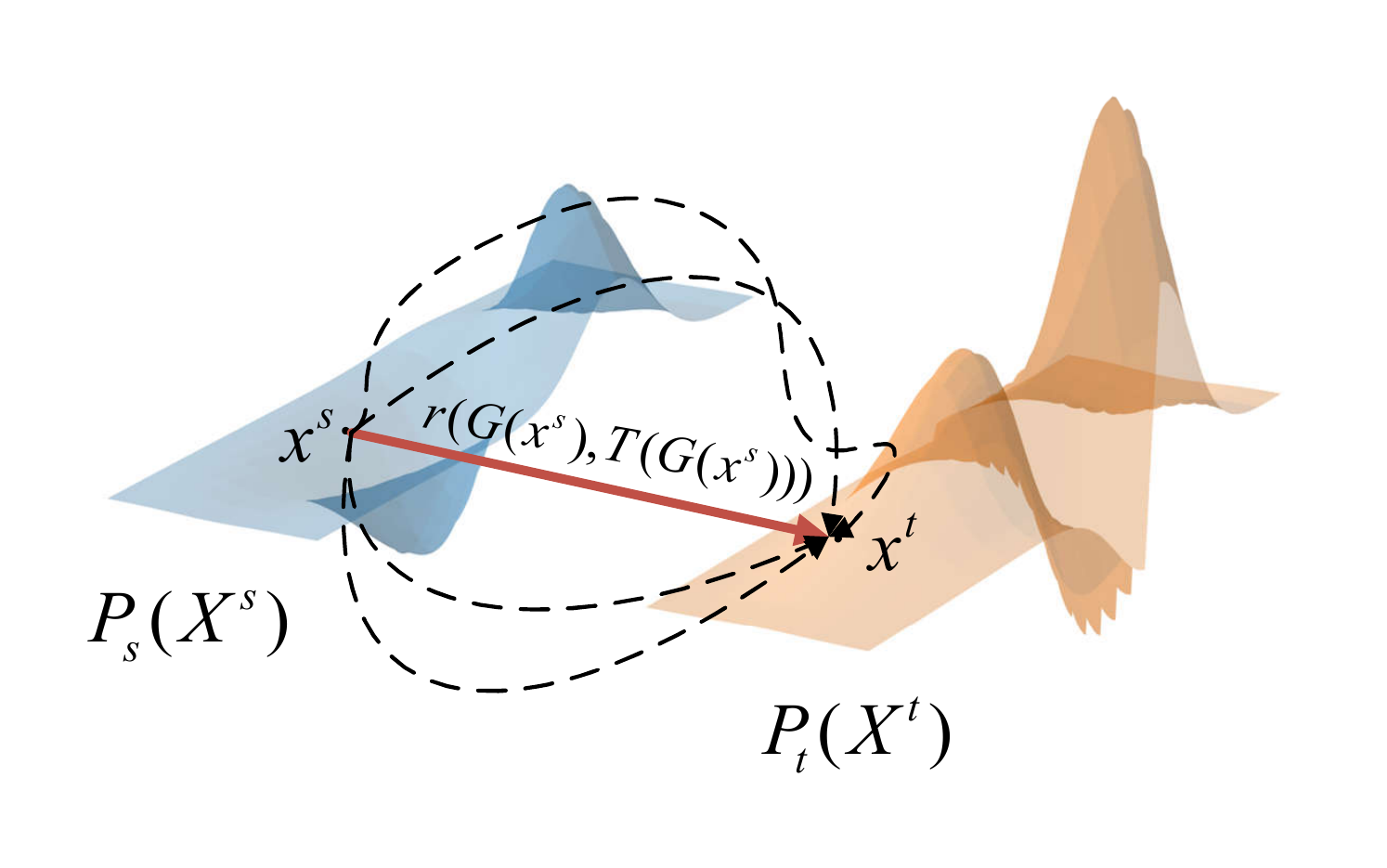}
\caption{When transferring features from the source to target domain, the regularization term proposed forces our model choosing the shortest path (red line).}
\label{fig:reg}
\end{figure}

For a transport problem, there are usually a few practical paths as shown in Fig.~\ref{fig:reg}. The optimal transport theory seeks the most efficient way of transforming one distribution of mass to another, just like the red line in Fig.~\ref{fig:reg}. In detail, $\int r({\bf x},T({\bf x}))d\mu({\bf x})$ in (\ref{eq7}), which indicates the expected cost of transportation, is to be minimized. If it reaches the minimum, the most efficient path would be found. Once we refine the optimal transport theory into unsupervised DA, the regularization term $r(G({\bf x}^s),T(G({\bf x}^s)))$ leads our model to select the most efficient way of transforming a source to target distribution. Specifically, the term attempts to constrain the distance between the features before and after the transformation. If we regard the distance as the cost of transformation, similar to the cost of transportation, the term attempts to select the shortest path from a number of transfer paths that map a source to target distribution.

On one hand, when the distributions of source and target domains are totally different, domain classifier $D$ can so easily distinguish samples from different domains that ${\cal L}_s$ and ${\cal L}_t$ backpropagate very small gradients. In this situation, the regularization term $r(G({\bf x}^s),T(G({\bf x}^s)))$ can still provide gradients to the target mapping. On the other hand, when $D$ directs updating parameters, the term would constrain the range of updating to prevent features from changing too rapidly, because it constrains differences between the features before and after the transformation. Thus, the stability of a training procedure of the proposed model is guaranteed via such added regularization term.

Consequently, our objective function becomes
\begin{align}
\displaystyle{\cal L}(\theta_d,\theta_g,\theta_c, \theta_t)=&\frac{1}{n_s}\sum_{{\bf x}_i\in D_s}{\cal L}_c(C(T(G({\bf x}_i))),y_i)-\nonumber\\
\displaystyle&\frac{\lambda}{n}(\sum_{{\bf x}_i\in D_s}{\cal L}_s(D(T(G({\bf x}_i))),d_i^s)+\nonumber\\
\displaystyle&\sum_{{\bf x}_i\in D_t}{\cal L}_t(D(G({\bf x}_i)),d_i^t))+\nonumber\\
\displaystyle&\beta \cdot r(G({\bf x}^s),T(G({\bf x}^s)))
\end{align}
where $\beta$ denotes the coefficient parameter of a regularization term. In addition, the optimization problem is to find parameters $\hat{\theta}_g$, $\hat{\theta}_d$, $\hat{\theta}_c$ and $\hat{\theta}_t$, where $\hat{\theta}_c$, $\hat{\theta}_g$, $\hat{\theta}_d$ and $\hat{\theta}_t$ satisfy:
\begin{align}
\hat{\theta}_d=& \argmin_{\theta_d}({\cal L}_s(\theta_d,\theta_g,\theta_t)+{\cal L}_t(\theta_d, \theta_g))\\
\hat{\theta}_c=& \argmin_{\theta_c}{\cal L}_c(\theta_g,\theta_c,\theta_t)\\
\hat{\theta}_g=& \argmin_{\theta_g}{\cal L}(\theta_d, \theta_g,\theta_c,\theta_t)\\
\hat{\theta}_t=& \argmin_{\theta_t}({\cal L}_c(\theta_g,\theta_c,\theta_t)+{\cal L}_s(\theta_d,\theta_g, \theta_t)+r(\theta_g, \theta_t))
\end{align}

In this case,~(\ref{eq12}) is reformulated into:
\begin{align}
\min_{G,T}{\cal L}(D^*,G,T)=2\cdot JSD(P_s||P_t)-2log2+r(\theta_g, \theta_t)
\end{align}
where $\frac{\partial r(\theta_g, \theta_t)}{\theta_g}$ and $\frac{\partial r(\theta_g, \theta_t)}{\theta_t}$ would not be zeros. Since, $JSD(P_s||P_t)=0$ appears easily, this regularization term provides gradients for parameters in $G$ and $T$, thereby alleviating the adverse effects of the vanishing gradient problem.

\subsection{Framework of Proposed Method}

In a training period for our model, we have two stages. In the  first one, feature extractor $G$ and transform network $T$ receive labeled source samples from $D_s\{{\bf x}^s_i,y^s_i,d^s_i\}_{i=1}^{n_s}$, and outputs ${\bf f}^s$ and $T({\bf f}^s)$. With class labels $y^s$ and domain labels $d^s$, ${\cal L}_c$ is computed by label classifier $C$, and ${\cal L}_s$ is computed by domain classifier $D$. At the same time, the regularization term $r(G({\bf x}^s),T(G({\bf x}^s)))$ is also obtained according to  ${\bf f}^s$ and $T({\bf f}^s)$. In the second stage, $G$ receives unlabeled samples from $D_t\{{\bf x}^t_i,d^s_i\}_{i=1}^{n_t}$, and outputs ${\bf f}^t$. Similarly, ${\cal L}_t$ is computed by domain classifier $D$. At last, all the above losses are multiplied by their corresponding coefficients, and then the model is optimized using these losses.

As for optimizing adversarial networks, previous studies have carried out a number of explorations~\cite{ganin2016domain,tzeng2017adversarial}. In~\cite{tzeng2017adversarial}, an iterative optimization strategy is proposed, where a feature extractor and domain classifier update their parameters iteratively. Specifically, it alternates between $k$ steps of optimizing a domain classifier and one step of optimizing a feature extractor. This is the most common training strategy which is also widely used in GANs~\cite{goodfellow2014generative,wang2017generative}. One of the obstacles to it is that tuning the hyperparameter $k$. Unsuitable $k$ may cause a failure of model training. As a result we have to tune this hyperparameter for each model carefully. Instead, in~\cite{ganin2016domain}, the proposed gradient reversal layer (GRL) replaces iterative optimization. During forward propagation, GRL has no difference from normal layers, whereas during backpropagation, GRL reverses the gradient from the subsequent layer, multiplies it by a coefficient $\gamma$ and passes it to the previous layer. Based on a large number of experiments,~\cite{ganin2016domain} adjusts $\gamma$ using the following formula: $\gamma=\frac{2}{1+e^{-10p}}-1$, where $p$ is the training progress linearly changing from 0 to 1. In the implementation of ARTN, we choose GRL to optimize our model. With this strategy, there is no need to tune the hyperparameter $k$, and parameters of the feature extractor and domain classifier are updated in one backpropagation.

Algorithm~\ref{alg:alg1} provides the pseudo-code of our proposed learning procedure. With stochastic gradient descent (SGD), parameters $\theta_d$, $\theta_c$ and $\theta_g$ are updated. When the loss converges, the training stops.

\begin{algorithm}[!h]
\caption{Learning Procedure of ARTN}
\label{alg:alg1}
\begin{algorithmic}[1]
\REQUIRE~~\\
Labeled source samples $D_s\{{\bf x}^s_i,y^s_i,d^s_i\}_{i=1}^{n_s}$\\
Unlabeled target samples $D_t\{{\bf x}^t_i,d^s_i\}_{i=1}^{n_t}$\\
Learning rate $\alpha$, Coefficient parameters $\lambda, \beta$
\ENSURE~~\\
Model parameters \{$\theta_d$,$\theta_g$,$\theta_c$,$\theta_t$\}
\WHILE {not converge} 
\FOR {$i$ from 1 to $n_s$}
\STATE ${\bf f}_i^s=G({\bf x}^s_i)$
\STATE ${\cal L}_c=crossentropy(C(T({\bf f}_i^s)),y^s_i)$
\STATE ${\cal L}_s=crossentropy(D(T({\bf f}_i^s)), d^s_i)$
\STATE $reg=r({\bf f}_i^s, T({\bf f}_i^s))$
\ENDFOR
\FOR {$i$ from 1 to $n_t$}
\STATE ${\bf f}_i^t=G({\bf x}^t_i)$
\STATE ${\cal L}_t=crossentropy(D({\bf f}_i^t), d^t_i)$
\ENDFOR
\STATE ${\cal L}_d\leftarrow {\cal L}_s+{\cal L}_t$
\STATE $\theta_d\leftarrow \theta_d - \alpha\cdot\frac{\partial {\cal L}_d}{\partial \theta_d}$
\STATE $\theta_c\leftarrow \theta_c - \alpha\cdot\frac{\partial {\cal L}_c}{\partial \theta_c}$
\STATE $\theta_g\leftarrow \theta_g - \alpha\cdot\frac{\partial ({\cal L}_c-\lambda {\cal L}_d + \beta reg)}{\partial \theta_g}$
\STATE $\theta_t\leftarrow \theta_t-\alpha\cdot\frac{\partial ({\cal L}_c-\lambda {\cal L}_s + \beta reg)}{\partial \theta_g}$
\ENDWHILE
\end{algorithmic}
\end{algorithm}

If $G$, $T$ and $D$ have enough capacity, and at each loop of Algorithm~\ref{alg:alg1}, $D$ is allowed to reach its optimal ${D^*}$ given $G$ and $T$, and $P_t$ is updated so as to improve the following criterion:
\begin{align}
{\mathbb{E}}_{{\bf x}\thicksim P_s}[logD^*(T(G({\bf x})))]+{\mathbb{E}}_{{\bf x}\thicksim P_t}[log(1-D^*(G({\bf x})))]
\end{align}
Then $P_t$ converges to $P_s$. Similar to~\cite{goodfellow2014generative}, we give a brief proof as follows.

\emph{Proof.} Consider that $U(P_t, D)=\int_{\bf z}P_s({\bf z})logD({\bf z})+P_t({\bf z})log(1-D({\bf z}))d{\bf z}$ as a function of $P_t$. Note that $U(P_t, D)$ is convex in $P_t$. The subderivatives of a supremum of convex functions include the derivative of the function at the point where the maximum is attained. In other words, if $f(x) =
sup_{\alpha\in\mathcal A} f_{\alpha}(x)$ and $f_{\alpha}(x)$ is convex in $x$ for every $\alpha$, then $\partial f_{\beta}(x)\in \partial f$ if $\beta = arg sup_{\alpha\in\mathcal A} f_{\alpha}(x)$.
This is equivalent to computing a gradient descent update for $P_t$ at $D^*$ given the corresponding
$G$ and $T$. $sup_{D} U(P_t, D)$ is convex in $P_t$ with a unique global optimum as proven in~(\ref{eq12}).
Hence, with sufficiently small updates of $P_t$, $P_t$ converges to $P_s$.

\section{Experiments}
\label{section4}

In order to evaluate the effectiveness of the proposed method, we test the proposed ARTN for unsupervised DA in several experiments that are recognized to be difficult. For the first experiment, we test our model  in a sentiment analysis task. Second, to test its performance when source and target domains are relatively similar, the model is evaluated on several digits datasets. Third, to test it when facing a large discrepancy between source and target domains, the model is evaluated on a natural image dataset. Fourth, to test its anti-noise and generalization abilities, we test it when target images are added with varying noise. Fifth, to test the effectiveness of regularization in the proposed method, we compare the performance of ARTN with and without regularization on a natural image dataset. Finally, we investigate the effects of parameter $\lambda$ on the performance of the proposed method. In all experiments, we implement models with Pytorch, and employ the learning strategy GRL mentioned in Section~\ref{section3}, which reverses and propagates gradients to the feature extractors.

\subsection{Sentiment Analysis}

We use the {\bf Amazon reviews} dataset with the same pre-processing used in mSDA~\cite{Chen:2012:MDA:3042573.3042781} and DANN~\cite{ganin2016domain}. It contains reviews of four different categories of products: {\tt Books}, {\tt DVDs}, {\tt Kitchen Appliances} and {\tt Electronics}, which means that this dataset includes four domains and we can set up twelve domain adaptation tasks across them. Reviews are encoded in 5 000 dimensional feature vectors of unigrams and bigrams, and labels are binary: 0 if a product is ranked up to 3 stars, and 1 if it is ranked 4 or 5 stars. In all twelve tasks, we use 2000 labeled source samples and 2000 unlabeled target samples to train our model. In a testing period, we test our model on separate target test sets (between 3000 and 6000 examples). To evaluate the effectiveness of our model, we compare it with DANN~\cite{ganin2016domain}, DAN~\cite{long2015learning}, Central Moment Discrepancy (CMD) for Domain-Invariant Representation Learning~\cite{DBLP:journals/corr/ZellingerGLNS17}, Variational Fair Autoencoder (VFAE)~\cite{louizos2015variational} and the model with no adaptation. The results are directly cited from the original pulication~\cite{ganin2016domain}.

In this experiment, we use the same neural network as DANN~\cite{ganin2016domain}. Both domain and label classifiers consist of just one layer with 100 hidden units followed by the final output layer. Because there is only one hidden layer in the neural network, we build just one residual connection. ReLU activation function and batch normalization are employed. We choose SGD as the optimizer with its learning rate 0.001 and momentum 0.9. Parameters $\lambda$ is set to 0.5, and $\beta$ is set to 0.1. The batch size is set to 128. All the results are recorded after convergence.

Results are shown in Table~\ref{tab:table3}. The accuracy of ARTN is the highest in three out of twelve domain adaptation tasks. The accuracy of CMD-based model is the highest in six tasks and VFAE achieves the highest accuracy in three tasks. Therefore, in the experiment of sentiment analysis, ARTN is comparable with VFAE and slightly worse than CMD.

\begin{table*}[htbp]
\centering
\caption{Classification accuracy percentage of sentiment analysis experiment among all twelve tasks. The first column corresponds to the performance if no adaption is implemented. The proposed method outperforms the others in three of twelve tasks.}
\label{tab:table3}
\begin{center}
\begin{small}
\begin{sc}
\begin{tabular}{lcccccc}
\hline\hline
Source$\rightarrow$Target & SOURCE ONLY& DAN~\cite{long2015learning} & DANN~\cite{ganin2016domain}&ARTN&CMD~\cite{DBLP:journals/corr/ZellingerGLNS17}&VFAE~\cite{louizos2015variational}\\
\hline
books$\rightarrow$dvd&78.7&79.6 &78.4&{\bf 81.4}&80.5&79.9\\
books$\rightarrow$electronics&71.4&75.8 &73.3& 77.5&78.7&{\bf 79.2}\\
books$\rightarrow$kitchen&74.5&78,7 &77.9&78.8&81.3&{\bf 81.6}\\
dvd$\rightarrow$books&74.6&78.0 &72.3&78.8&{\bf 79.5}&75.5\\
dvd$\rightarrow$electronics&72.4&76.6 &75.4&77.0&{\bf 79.7}&78.6\\
dvd$\rightarrow$kitchen&76.5&79.6 &78.3&79.3&{\bf 83.0}&82.2\\
electronics$\rightarrow$books&71.1&73.3 &71.3&72.4&{\bf 74.4}&72.7\\
electronics$\rightarrow$dvd&71.9&74.8 &73.8&73.9& 76.3&{\bf 76.5} \\
electronics$\rightarrow$kitchen&84.4&85.7 &85.4 &{\bf 86.4}&86.0&85.0 \\
kitchen$\rightarrow$books&69.9&74.0 &70.9 &73.8&{\bf 75.6}&72.0 \\
kitchen$\rightarrow$dvd&73.4&76.3 &74.0 & 75.7&{\bf 77.5}&73.3\\
kitchen$\rightarrow$electronics&83.3&84.4 &84.3 &{\bf 86.1}&85.4&83.8 \\
\hline\hline
\end{tabular}
\end{sc}
\end{small}
\end{center}
\end{table*}

\subsection{Digits}

\begin{figure}[htp]
\centering
\includegraphics[width = \columnwidth]{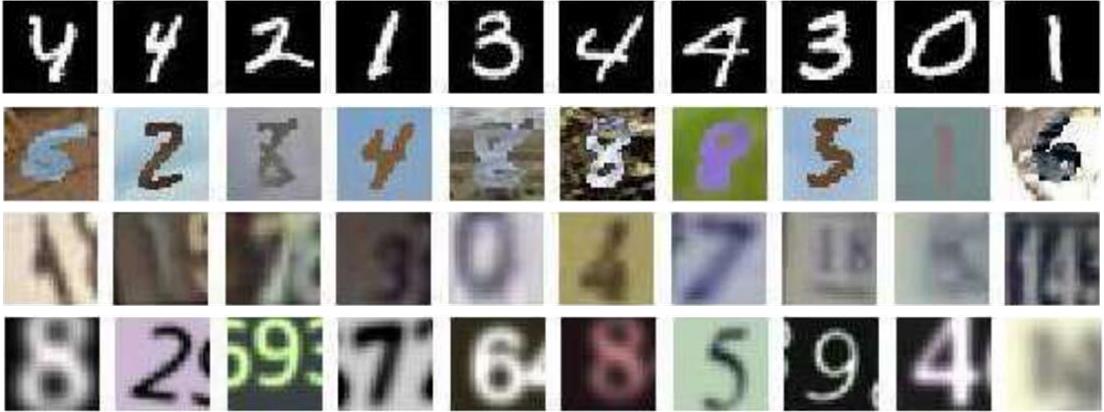}
\caption{Samples of digits dataset. The first to last rows  correspond to MNIST, MNIST-M, SVHN and SYN NUMS.}
\label{fig:digits}
\end{figure}

In order to evaluate the performance when the discrepancy between source and target domains is relatively small, we experimentally test ARTN in several pairs of unsupervised domain adaptation tasks whose images are from the {\bf MNIST}, {\bf MNIST-M}, {\bf SVHN} and {\bf SYN NUMS} digits datasets. All these datasets consist of 10 classes, and we use the full training sets in all datasets. Example images from each dataset are shown in Fig.~\ref{fig:digits}. In this experiment, we set three transfer tasks: MNIST$\rightarrow$MNIST-M, SVHN$\rightarrow$MNIST, and SYN NUMS$\rightarrow$SVHN. As is shown in Fig.~\ref{fig:digits}, images in SYN NUMS and SVHN are similar, whereas images in MNIST are much different from the other digits datasets.

We choose several unsupervised DA approaches to compare with the proposed one. CORAL~\cite{sun2016return}, CMD~\cite{DBLP:journals/corr/ZellingerGLNS17} and DAN~\cite{long2015learning} rely on the distance metric between source and target distributions. DANN~\cite{ganin2016domain}, CoGAN~\cite{liu2016coupled}, Domain Transfer Network (DTN)~\cite{taigman2016unsupervised}, CYCADA~\cite{pmlr-v80-hoffman18a} and ADDA~\cite{tzeng2017adversarial} are based on adversarial learning. 

For MNIST$\rightarrow$MNIST-M, we use a simple modified LeNet~\cite{lecun1998gradient}. As for a domain classifier, we stack two fully connected layers: one layer with 100 hidden units followed by the final output layer. Each hidden unit uses a ReLU activation function. For SVHN$\rightarrow$MNIST and SYN NUMS$\rightarrow$SVHN, we use a three-layer convolutional network as a feature extractor, and a three-layer fully connected network as a domain classifier. In all tasks, batch normalization is employed. We employ SGD with 0.01 learning rate and the momentum 0.9. $\lambda$ is set to 1, and $\beta$ is set to 0.2. The batch size is set to 128. Prediction accuracy in the target domain is reported after convergence. 

Results of our digits experiment are shown in Table~\ref{tab:table1}. Note that the basic networks for DTN and CYCADA are different from others, and the accuracy with no adaptation is included in the bracket. In MNIST$\rightarrow$MNIST-M, the proposed model's accuracy is 85.6\% which outperforms the best of other methods by 0.6\%. In SYN NUMS$\rightarrow$SVHN, its accuracy achieves 89.1\%, which is comparable to DANN's. In SYN NUMS$\rightarrow$SVHN, the accuracy of ARTN is 85.8\%, which is just lower than CYCADA's. Note that, CYCADA achieves the higher accuracy with better basic networks. For a fair comparison, the improvements compared with the adaptation-free models of ARTN, DTN, CyCADA pixel only and CyCADA pixel+feat are 30.9\%, 8.3\%, 3.2\% and 23.3\%, respectively. It is obvious that ARTN brings a bigger boost to the adaptation-free model. Totally, in two of three tasks, our approach outperforms other methods, and in the task whose source and target datasets are similar, it can achieve the same competitive results as the others.

\begin{table*}[htbp]
\centering
\caption{Classification accuracy percentage of digits classifications among MNIST, MNIST-M, SVHN and SYN NUMS. The first row corresponds to the performance if no adaption is implemented. The proposed method outperforms the others in two of three tasks when it comes to improvement compared with the basic network. In addtiont, the results are cited from literature.
}
\label{tab:table1}
\begin{center}
\begin{small}
\begin{sc}
\begin{tabular}{lccc}
\hline\hline
Method& MNIST$\rightarrow$MNIST-M& SYN NUMS$\rightarrow$SVHN& SVHN$\rightarrow$MNIST\\
\hline
Source only&51.4&86.7&54.9\\
CORAL~\cite{sun2016return}&57.7&85.2&63.1\\
DAN~\cite{long2015learning}&76.9&88.0&71.1\\
DANN~\cite{ganin2016domain}&76.7&{\bf 91.1}&73.9\\
CMD~\cite{DBLP:journals/corr/ZellingerGLNS17}&85.0&85.5&84.5\\
CoGAN~\cite{liu2016coupled}&-&-&diverge\\
ADDA~\cite{tzeng2017adversarial}&-&-&76.0\\
DTN~\cite{taigman2016unsupervised}&-&-&84.4(76.1)\\
CyCADA pixel only~\cite{pmlr-v80-hoffman18a}&-&-&70.3(67.1)\\
CyCADA pixel+feat~\cite{pmlr-v80-hoffman18a}&-&-&{\bf 90.4}(67.1)\\
ARTN&{\bf 85.6}&89.1&85.8\\
\hline\hline
\end{tabular}
\end{sc}
\end{small}
\end{center}
\end{table*}

\begin{table*}[!t]
\renewcommand{\arraystretch}{1.3}
\centering
\caption{Classification accuracy percentage of experiment on the Office-31 dataset. The first column corresponds to the performance if no adaption is implemented. The second to last columns correspond to the performance of different DA methods and the proposed method.}
\label{tab:table2}
\begin{tabular}{lcccc}
\hline\hline
Method & DSLR$\rightarrow$AMAZON & WEBCAM$\rightarrow$AMAZON & AMAZON$\rightarrow$WEBCAM & AMAZON$\rightarrow$DSLR\\
\hline
AlexNet&51.1&49.8&61.6&63.8\\
DDC~\cite{tzeng2015simultaneous}&52.1&52.2&61.8&64.4\\
Deep CORAL~\cite{sun2016return}&52.8&51.5&66.4&66.8\\
DAN~\cite{long2015learning}&54.0&53.1&68.5&67.0\\
\hline
InceptionBN&60.1&57.9&70.3&70.5\\
LSSA~\cite{aljundi2015landmarks}&57.8&57.8&67.7&71.3\\
CORAL~\cite{sun2016return}&59.0&60.2&70.9&71.9\\
AdaBN~\cite{li2016revisiting}&59.8&57.4&74.2&73.1\\
\hline
VGG16&58.2&57.8&67.6&73.9\\
CMD~\cite{DBLP:journals/corr/ZellingerGLNS17}&{\bf 63.8}&{\bf 63.3}&{\bf 77.0}&{\bf 79.6}\\
\hline
ResNet34 & 57.5 & 55.5 & 68.4 & 68.9 \\
DANN~\cite{ganin2016domain} & 58.1 & 56.3 & 73.7 & 75.3\\
ARTN & 60.9 & 61.0 & 76.2 & 76.1 \\
\hline\hline
\end{tabular}
\end{table*}

\subsection{Image Classification}

\begin{figure}[htp]
\centering
\includegraphics[width = \columnwidth]{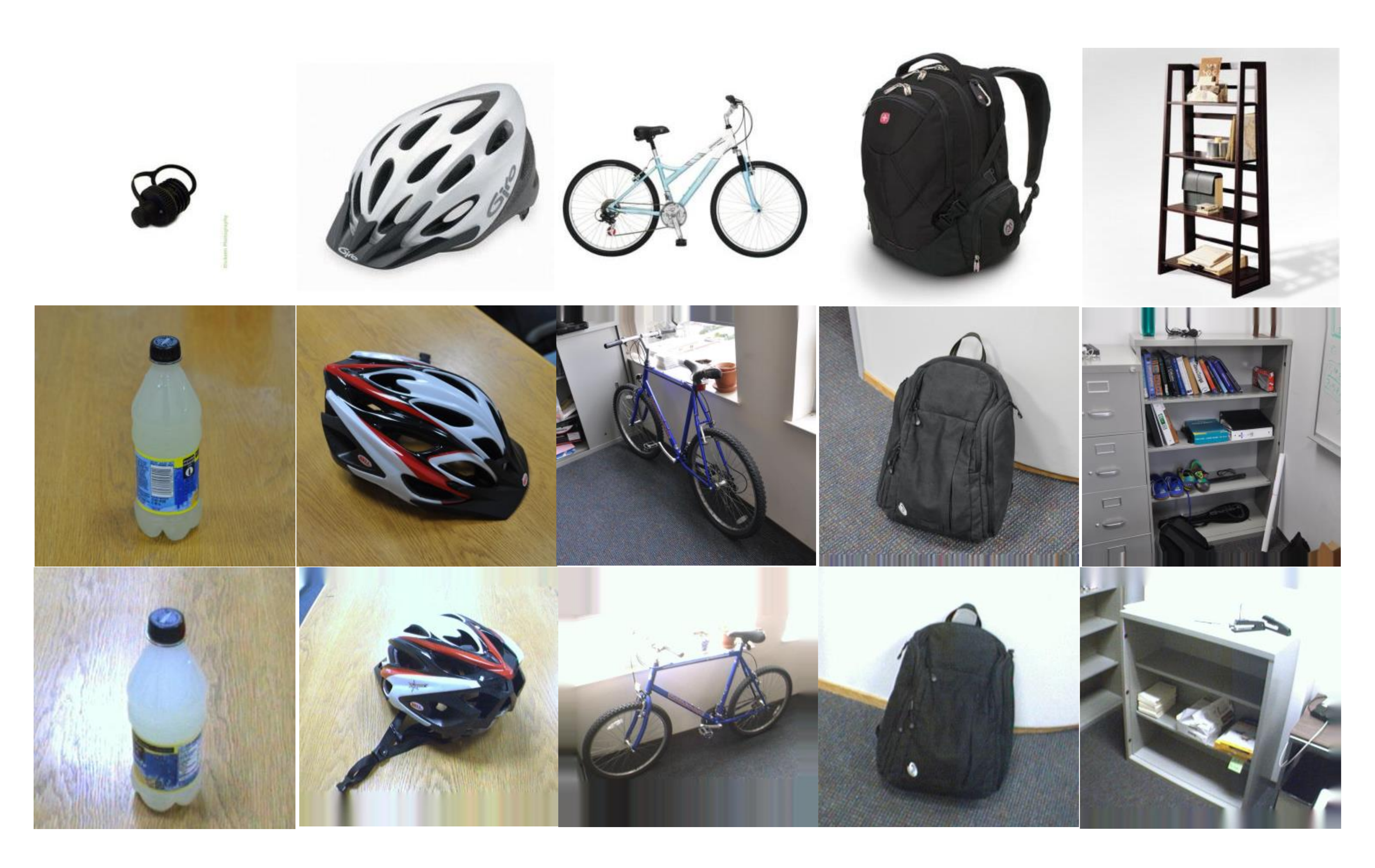}
\caption{Samples of Office-31 dataset. The first to last rows  correspond to AMAZON, DSLR and WEBCAM.}
\label{fig:office}
\end{figure}

We further evaluate our model in a more complex setting. The proposed model is tested on a natural image dataset {\bf Office-31}, which is a standard benchmark for visual domain adaptation, comprising 4,110 images and 31 categories collected from three domains: {\tt AMAZON} ({\bf A}, images downloaded from amazon.com) with 2,817 images, {\tt DSLR} ({\bf D}, high-resolution images captured by a digital SLR camera) with 498 images and {\tt WEBCAM} ({\bf W}, low-resolution images captured by a Web camera) with 795 images. Samples of {\bf Offfice-31} dataset are shown in Fig.~\ref{fig:office}. In order to test the generalization ability of different methods, we focus on the most difficult four tasks~\cite{long2015learning}: {\bf A}$\rightarrow${\bf D}, {\bf A}$\rightarrow${\bf W}, {\bf D}$\rightarrow${\bf A} and {\bf W}$\rightarrow${\bf A}. In {\bf A}$\rightarrow${\bf W} and {\bf A}$\rightarrow${\bf D}, models are easier to train because images in source domain {\bf A} are adequate. In {\bf W}$\rightarrow${\bf A} and {\bf D}$\rightarrow${\bf A}, there are only hundreds of images in the source domain but about 2,900 images in the target one. Thus models are very difficult to train. In addition, we test our model without regularization to analyze how the regularization term of our model affects its performance. In this experiment, we evaluate the effectiveness of our approach by comparing it with different models trained on the {\bf Office-31} dataset. Note that some of the methods, such as DDC~\cite{tzeng2015simultaneous}, Deep CORAL~\cite{sun2016return} and DAN~\cite{long2015learning}, are based on AlexNet, some of the methods, such as LSSA~\cite{aljundi2015landmarks}, CORAL~\cite{sun2016return} and AdaBN~\cite{li2016revisiting}, are based on InceptionBN, and CMD~\cite{DBLP:journals/corr/ZellingerGLNS17} is based on VGG16. Results of these methods are cited from original papers. Moreover, we implement DANN~\cite{ganin2016domain} and the model with no adaptation to be the baselines.

Because of lacking sufficient images, we implement our model based on ResNet34~\cite{He_2016_CVPR} which is pre-trained on an ImageNet dataset, and fine-tune the model on Office-31. Different from the digits experiment, we build a residual connection for every three layers in ResNet34 instead of every layer. As for the domain classifier, we use a network with three fully connected layers. In addition, we replace the last layer of ResNet34 with a three-layer fully connected network, and use it to predict the labels of inputs. In all tasks, we employ the same SGD and parameter setting as before except that $\lambda$ is set to 0.6 and batch size is 40. All the prediction accuracy results are recorded after training for 30 epochs.

Results of the experiment on Office-31 are shown in Table~\ref{tab:table2}. In {\bf D}$\rightarrow${\bf A}, {\bf W}$\rightarrow${\bf A}, {\bf A}$\rightarrow${\bf W} and {\bf A}$\rightarrow${\bf D}, the proposed model achieves the accuracy of 60.9\%, 61.0\%, 76.2\% and 76.1\%, respectively. 
Thus, in all four tasks, the proposed model achieves the second highest accuracy. Note that these methods are based on different basic networks, besides the accuracy, improvement compared with the corresponding basic network is a fairer metric. The improvements of ARTN in four tasks are 3.4\%, 5.5\%, 7.8\% and 7.2\%, respectively. CMD, which outperforms all the related state-of-the-art methods on all four tasks, achieves 5.6\%, 5.5\%, 9.4\% and 5.7\% improvements. ARTN achieves a higher improvement in {\bf A}$\rightarrow${\bf D} and same improvement in {\bf W}$\rightarrow${\bf A} compared with the state-of-art method, CMD. The intuitive interpretation of CMD's excellent performance is that it minimizes the sum of differences of higher order central moments of the corresponding activation distributions. Higher-order statistics describe the differences between distributions more comprehensively, but they also incur significantly more computational overhead than such methods as our proposed one. In practice, the number of moments is pre-set to be no more than five. In summary, ARTN outperforms all the other methods except CMD.

\subsection{Generalization Analysis}

\begin{figure}
  \centering
  \subfigure[MNIST$\rightarrow$MNIST-M]{
    \label{fig:subfig:a} 
    \includegraphics[width=\columnwidth]{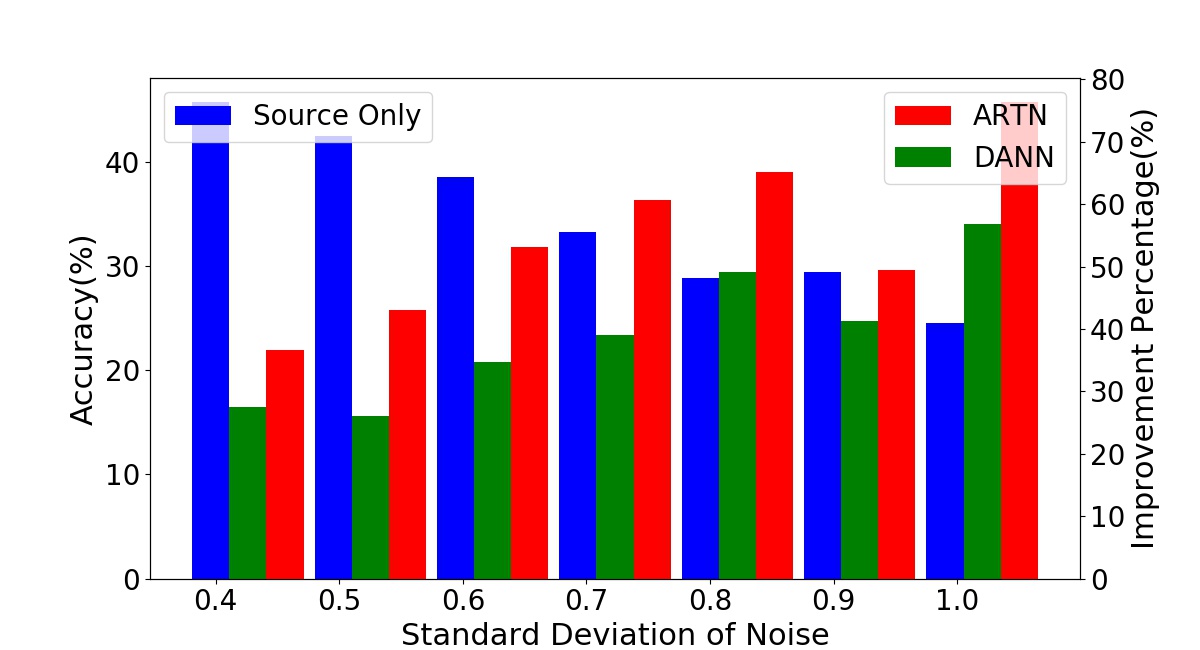}}
  \hspace{1in}
  \subfigure[SYN NUMS$\rightarrow$SVHN]{
    \label{fig:subfig:b} 
    \includegraphics[width=\columnwidth]{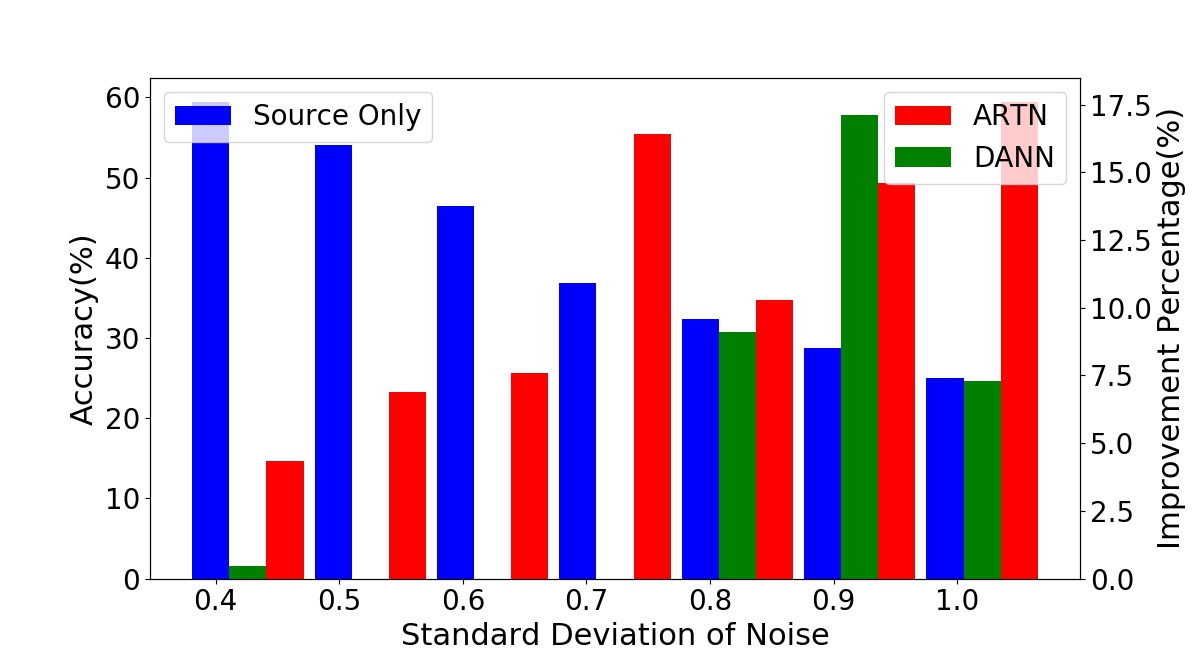}}
  \hspace{1in}
  \subfigure[SVHN$\rightarrow$MNIST]{
    \label{fig:subfig:c} 
    \includegraphics[width=\columnwidth]{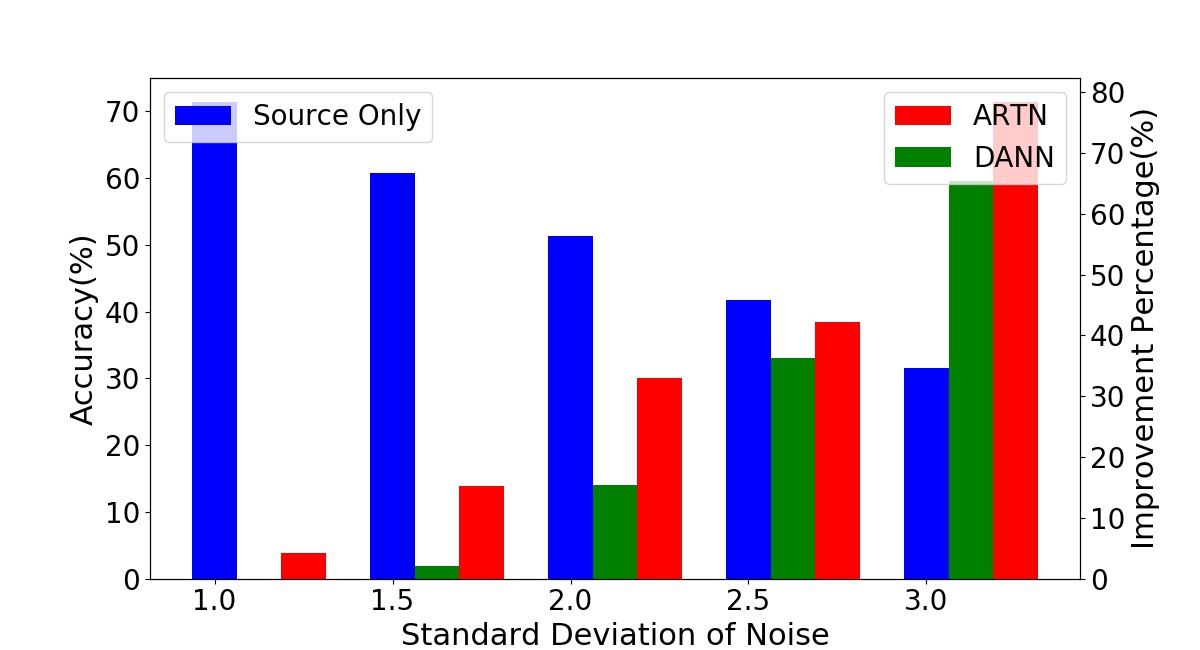}}
  \caption{Accuracy of the adaptation-free method and improvement of DANN and our model in MNIST$\rightarrow$MNIST-M, SYN NUMS$\rightarrow$SVHN and SVHN$\rightarrow$MNIST, where we add gaussian noise to images in the target domain. X-axis represents the standard deviation of noise, and Y-axis represents the accuracy of adaptation-free method and improvement percentage of DANN and our model in the target domain.}
  \label{fig:noise} 
\end{figure}

A generalization test is taken by adding Gaussian noise to images in a target domain. In this way, the discrepancy between source and target domains is larger and discriminative information in a target domain is more difficult to capture. In this experiment, we test the anti-noise and generalization abilities of our model based on the digits experiment. For images in the source domain, we follow the settings in MNIST$\rightarrow$MNIST-M, SYN NUMS$\rightarrow$SVHN and SVHN$\rightarrow$MNIST respectively, however, for images in the target domain, we add varying Gaussian noise. For MNIST$\rightarrow$MNIST-M and SYN NUMS$\rightarrow$SVHN, the standard deviation of Gaussian noise is selected from $\{0.4, 0.5, 0.6, 0.7, 0.8, 0.9, 1.0\}$. That in SVHN$\rightarrow$MNIST is from $\{1.0, 1.5, 2.0, 2.5, 3.0\}$. The means of Gaussian noise in all tasks are 0. Results are plotted in Fig.~\ref{fig:noise}. The baseline method is a model without adaptation. We also compare the proposed method with DANN~\cite{ganin2016domain}.

Comparing the proposed model with the adaptation-free model, we can see that although noises are added to the test images, the proposed model exhibits a great advantage over the adaptation-free model. In MNIST$\rightarrow$MNIST-M, when the standard deviation is 0.4, the accuracy of the adaptation-free model is 45.83\%, whereas ours improves it by 36.6\%. When standard deviation is 1.0, the accuracy of the adaptation-free model is 24.55\%, whereas ours improves is by 76.4\%. Similar results appear in SYN NUMS$\rightarrow$SVHN and SVHN$\rightarrow$MNIST, where the rate of improvement generally shows an upward trend in the case of a gradual increase of noise. Therefore, as the discrepancy between source and target domains increases, the performance advantage of the proposed model is becoming more and more obvious in comparison with a adaptation-free model. At the same time, the improvement percentage of our model is higher than DANN in almost all tasks, which means that the proposed method has better anti-noise abilities than DANN. This result demonstrates that even if there exists noise in a target domain, the proposed model can maintain excellent generalization and anti-noise abilities.

\begin{figure}[htp]
\centering
\includegraphics[width = \columnwidth]{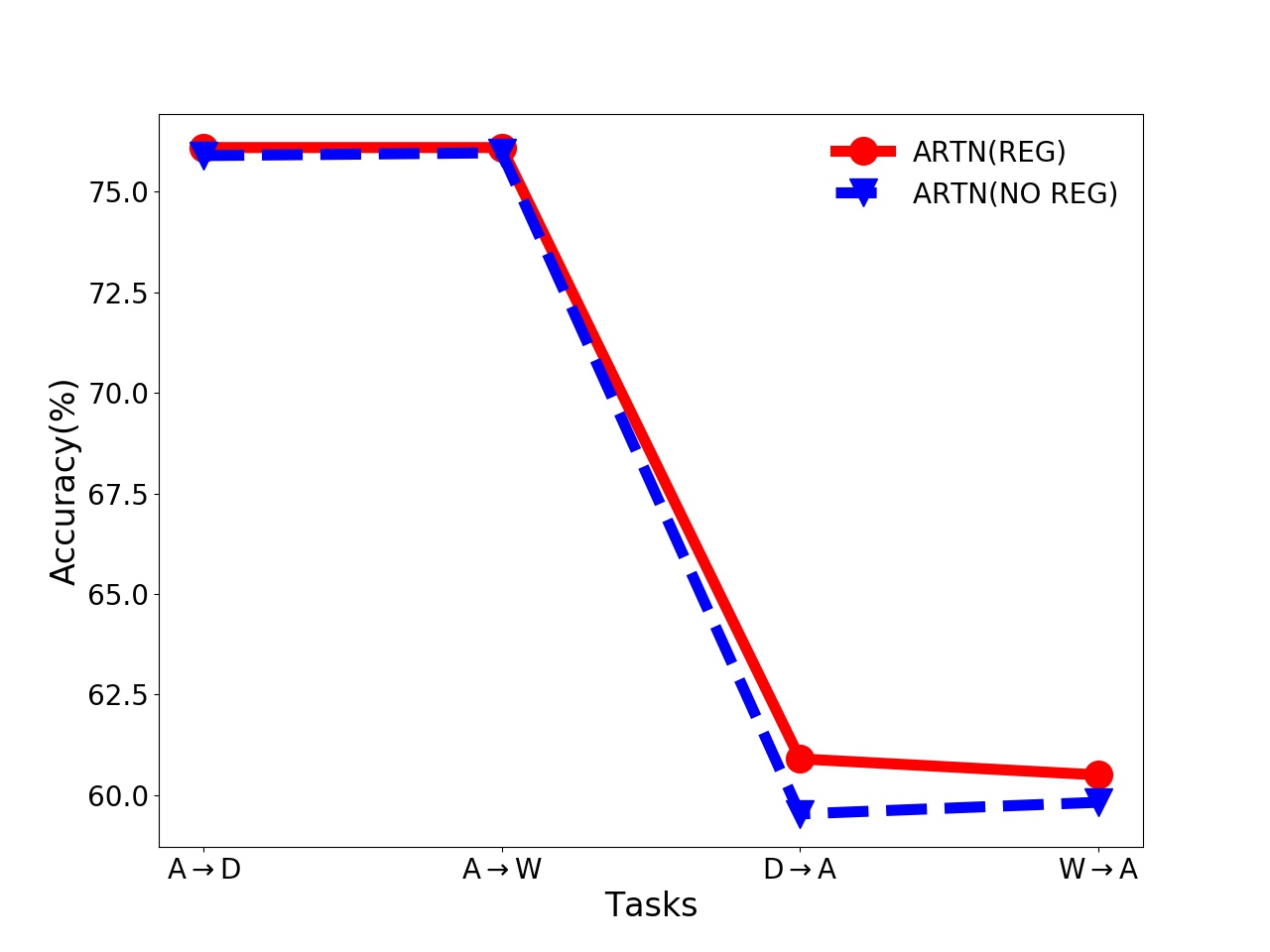}
\caption{Classification accuracy percentage of experiment on the Office-31 dataset.The red line corresponds to the proposed method with regularization and the blue line corresponds to the one without regularization. The regularizaton term shows a positive effect on the performance improvement.}
\label{fig:reg3}
\end{figure}

\subsection{Regularization Analysis}

We next analyze how the regularization term of our model affects the performance of our model. We test our model without regularization on Office-31 by setting $\beta =0$. In this way, ${\cal L}$ consists of ${\cal L}_c$, ${\cal L}_s$ and ${\cal L}_t$ only. Except for the regularization term, this experiment has same settings as the image classification experiment does. 

Results of this experiment are shown in Fig.~\ref{fig:reg3}. In {\bf D}$\rightarrow${\bf A}, {\bf W}$\rightarrow${\bf A}, {\bf A}$\rightarrow${\bf W} and {\bf A}$\rightarrow${\bf D}, the proposed model without regularization achieves the accuracy of 59.5\%, 59.8\%, 76.0\% and 75.9\%, which is lower by 1.4\%, 1.2\%, 0.2\% and 0.2\% of the proposed one with such term, respectively. The model with regularization outperforms DANN and the model without regularization in all tasks, which demonstrates the effectiveness of regularization. In another word, the regularization term strengthens the generalization ability of the proposed model. It should be noted that in {\bf D}$\rightarrow${\bf A}, {\bf W}$\rightarrow${\bf A} and {\bf A}$\rightarrow${\bf W}, the proposed model without regularization still outperforms DANN. This means that the improvement is not only from the regularization but also the modification of its architecture.

\begin{figure}
  \centering
  \subfigure[A$\rightarrow$W]{
    \label{fig:subfig:gAW} 
    \includegraphics[width=0.48\columnwidth]{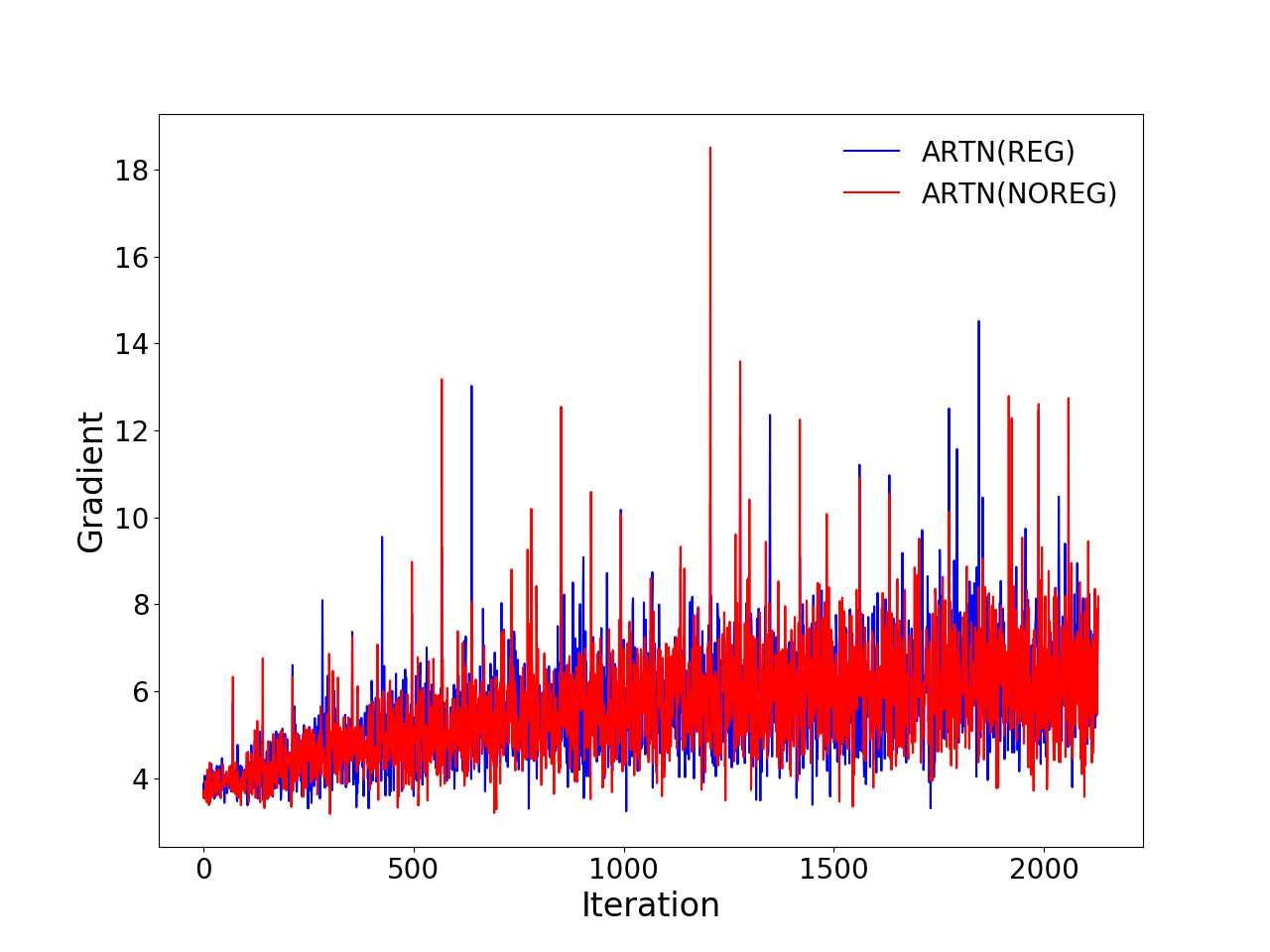}}
  \subfigure[A$\rightarrow$D]{
    \label{fig:subfig:gAD} 
    \includegraphics[width=0.48\columnwidth]{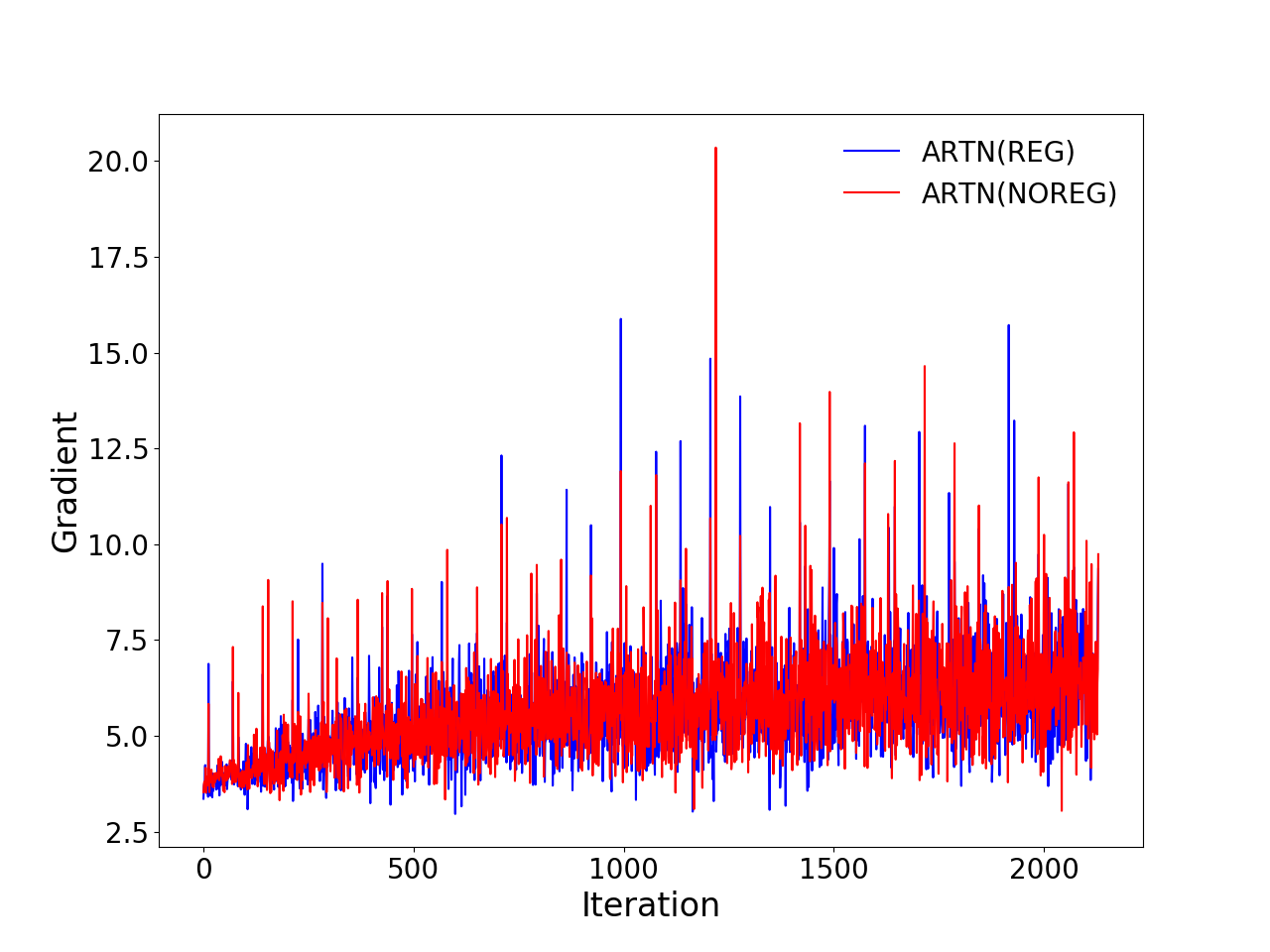}}
  \subfigure[W$\rightarrow$A]{
    \label{fig:subfig:gWA} 
    \includegraphics[width=0.48\columnwidth]{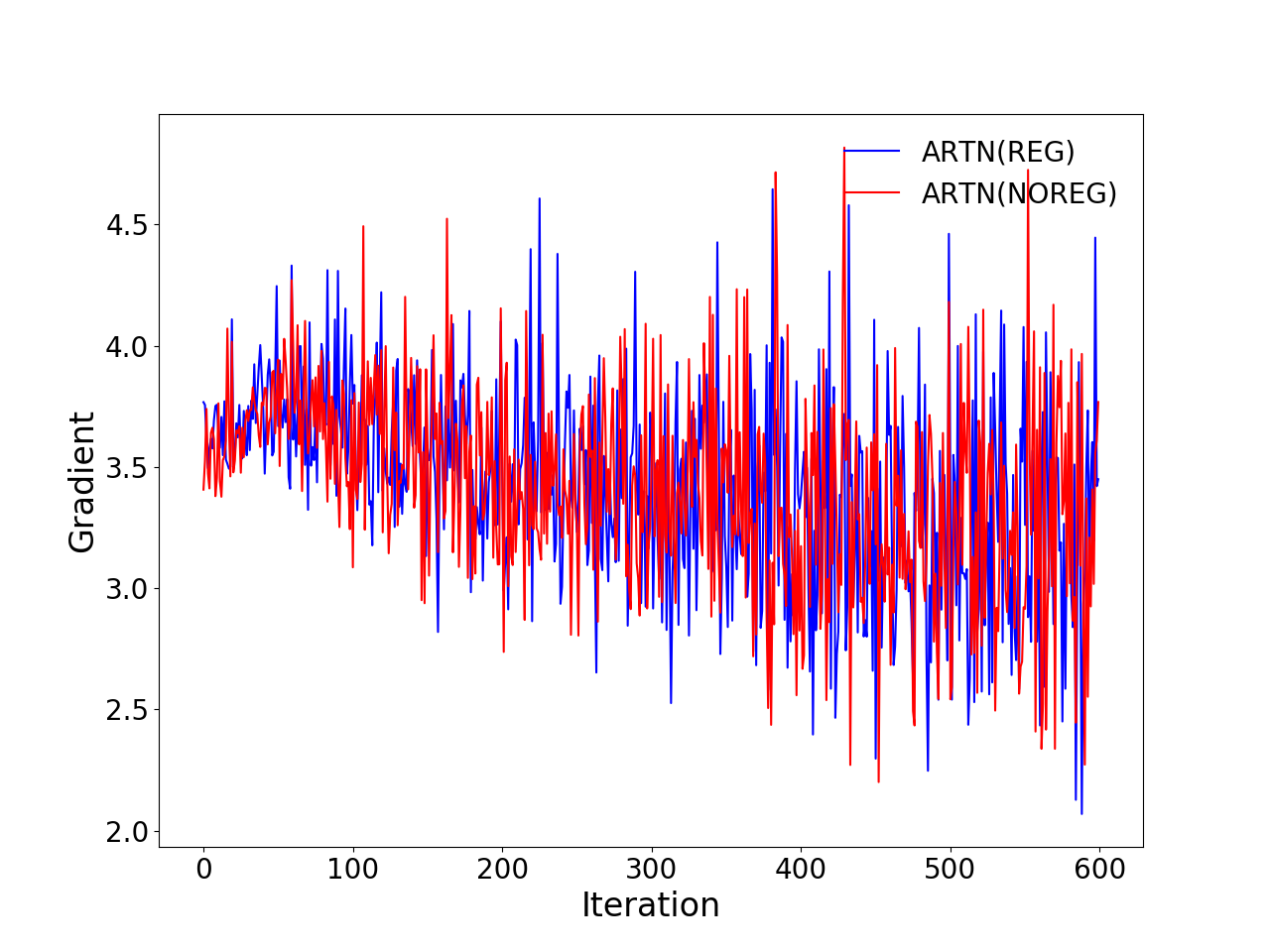}}
  \subfigure[D$\rightarrow$A]{
    \label{fig:subfig:gDA} 
    \includegraphics[width=0.48\columnwidth]{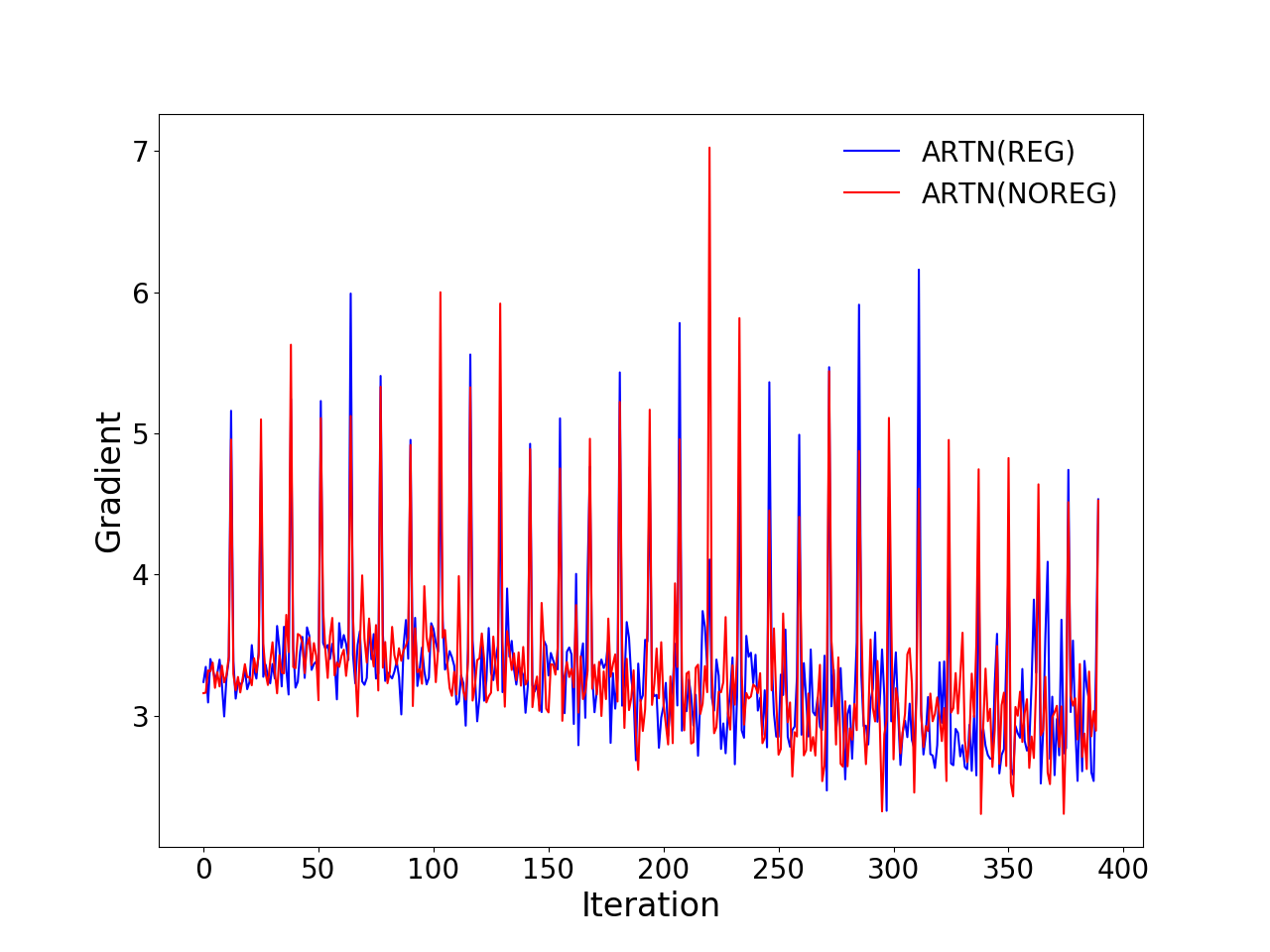}}  
\caption{$||\nabla_{\theta}{\cal L}(\theta)||$ in task {\bf D}$\rightarrow${\bf A}, {\bf W}$\rightarrow${\bf A}, {\bf A}$\rightarrow${\bf W} and {\bf A}$\rightarrow${\bf D}. Red line corresponds to our model without regularization term while blue line corresponds to our model with regularization term.}
  \label{fig:grad} 
\end{figure}

Besides performance improvement, we analyze how the regularization term affects the gradients during training. Because displaying every gradient of a parameter is impossible, we calculate $||\nabla_{\theta}{\cal L}(\theta)||$ to capture the overall statistics which is a metric adopted in~\cite{arjovsky2017wasserstein}. We record $||\nabla_{\theta}{\cal L}(\theta)||$ of the model with and without a regularization term on Office-31. Moreover, we record the minimum, maximum and standard deviation of $||\nabla_{\theta}{\cal L}(\theta)||$ during the training period. $||\nabla_{\theta}{\cal L}(\theta)||$ in {\bf D}$\rightarrow${\bf A}, {\bf W}$\rightarrow${\bf A}, {\bf A}$\rightarrow${\bf W} and {\bf A}$\rightarrow${\bf D} are drawn in Fig.~\ref{fig:grad} and related statistical data are shown in Table~\ref{tab:table4}. Please note that even if the gradient vanishing issue occurs during training, $||\nabla_{\theta}{\cal L}(\theta)||$ would not be very close to zero because it involves gradients of all parameters. 

According to Fig.~\ref{fig:grad}, especially in Fig.~\ref{fig:subfig:gAW}, (b) and (d), we can see that gradients of ARTN without a regularization term are easier to be unstable. There are more extreme large gradients in ARTN without a regularization term. The results are shown in Table~\ref{tab:table4}. In all four tasks, ARTN with a regularization term gets smaller standard deviation than ARTN without it. This directly indicates that a regularization term promotes the stability of adversarial training in our model. In detail, the maximum and minimum gradients are also recorded. We find that maximum gradient of ARTN with a regularization term in four tasks are smaller than that of ARTN without it. Meanwhile, except in {\bf W}$\rightarrow${\bf A}, the gap between maximum and minimum gradients suggests that ARTN with a regularization term is more stable than ARTN without it with a large margin. This fact can also be observed in Fig.~\ref{fig:grad}. Thus, the effect of the proposed regularization term to stabilize adversarial training in our model is considered being verified in this experiment. The reason why the {\bf W}$\rightarrow${\bf A} case is an exception needs to be explored as future research.

\begin{table}[!t]
\renewcommand{\arraystretch}{1.3}
\centering
\caption{Statistical data of $||\nabla_{\theta}{\cal L}(\theta)||$ are recorded. Both the models with and without regularization term are evaluated on Office-31. For each row, the upper line corresponds to ARTN with regularization and the lower line corresponds to ARTN without regularization.}
\label{tab:table4}
\begin{tabular}{lcccc}
\hline\hline
Method & A$\rightarrow$W & A$\rightarrow$D & W$\rightarrow$A & D$\rightarrow$A\\
\hline
\multirow{2}{*}{max}& 14.51& 15.88&4.64 & 6.16\\
& 18.50&20.35 &4.81 &7.02 \\
\hline
\multirow{2}{*}{min}&3.24 &2.96 &2.07 &2.32 \\
&3.17 &3.04 &2.20 &2.32 \\
\hline
\multirow{2}{*}{max-min}&11.27 &12.92 &2.57 &3.84 \\
&15.33 &17.31 &2.61 &4.70 \\
\hline
\multirow{2}{*}{std}&1.25 &1.36 &0.38 &0.59\\
&1.30 &1.37 &0.40 &0.61\\
\hline\hline
\end{tabular}
\end{table}

\begin{figure}
  \centering
  \subfigure[A$\rightarrow$W]{
    \label{fig:subfig:AW} 
    \includegraphics[width=0.48\columnwidth]{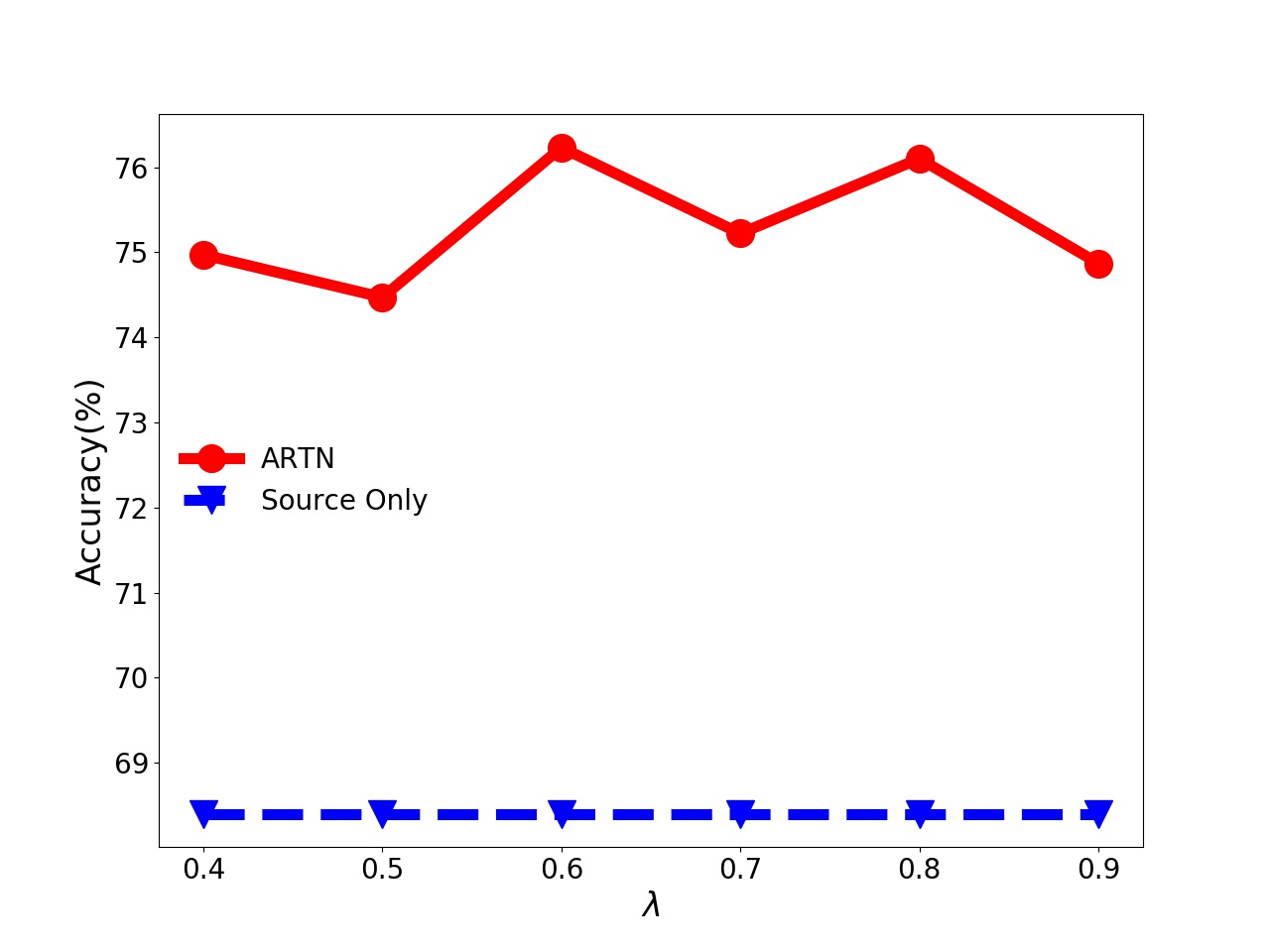}}
  \subfigure[A$\rightarrow$D]{
    \label{fig:subfig:AD} 
    \includegraphics[width=0.48\columnwidth]{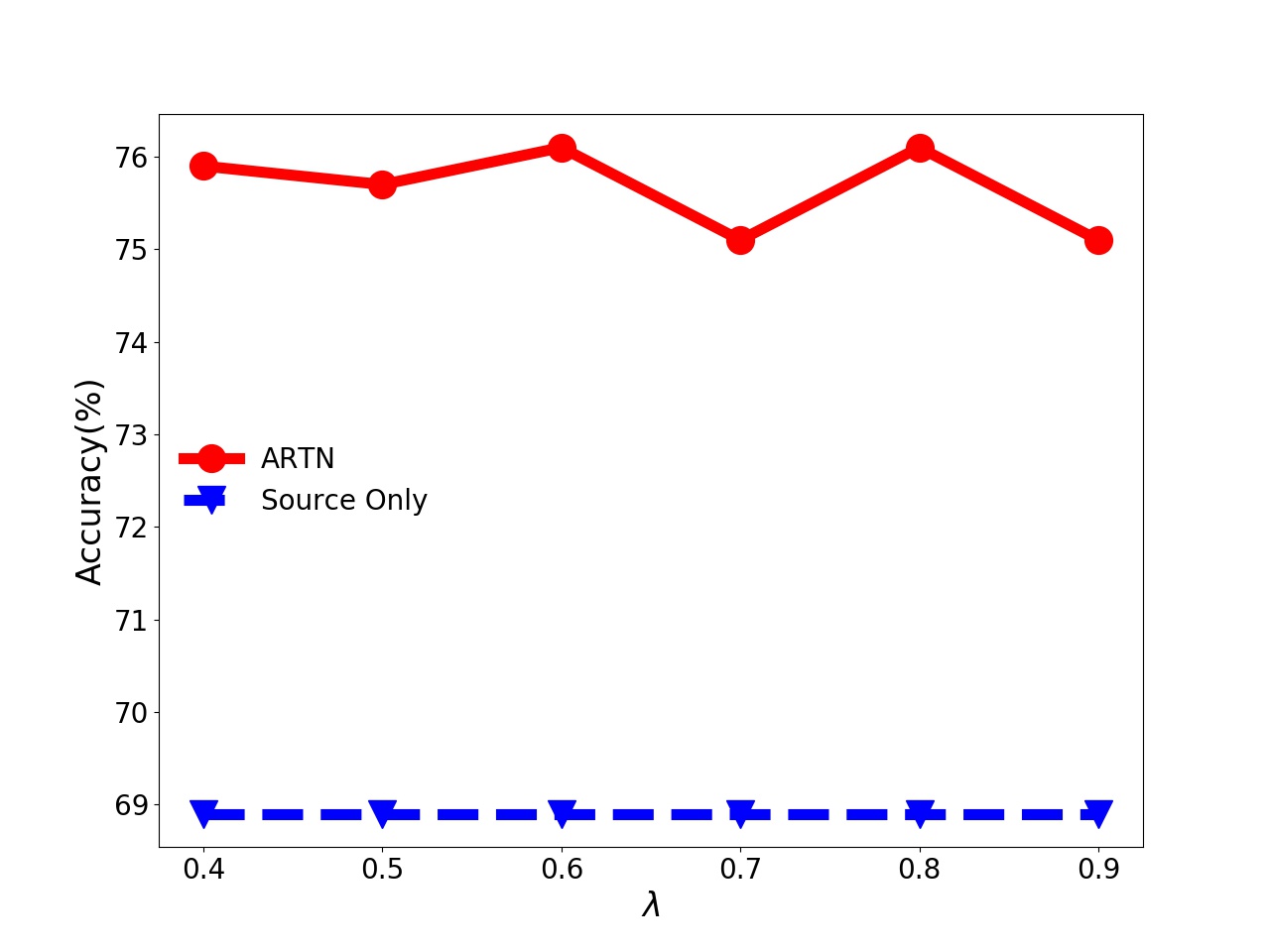}}
  \subfigure[W$\rightarrow$A]{
    \label{fig:subfig:WA} 
    \includegraphics[width=0.48\columnwidth]{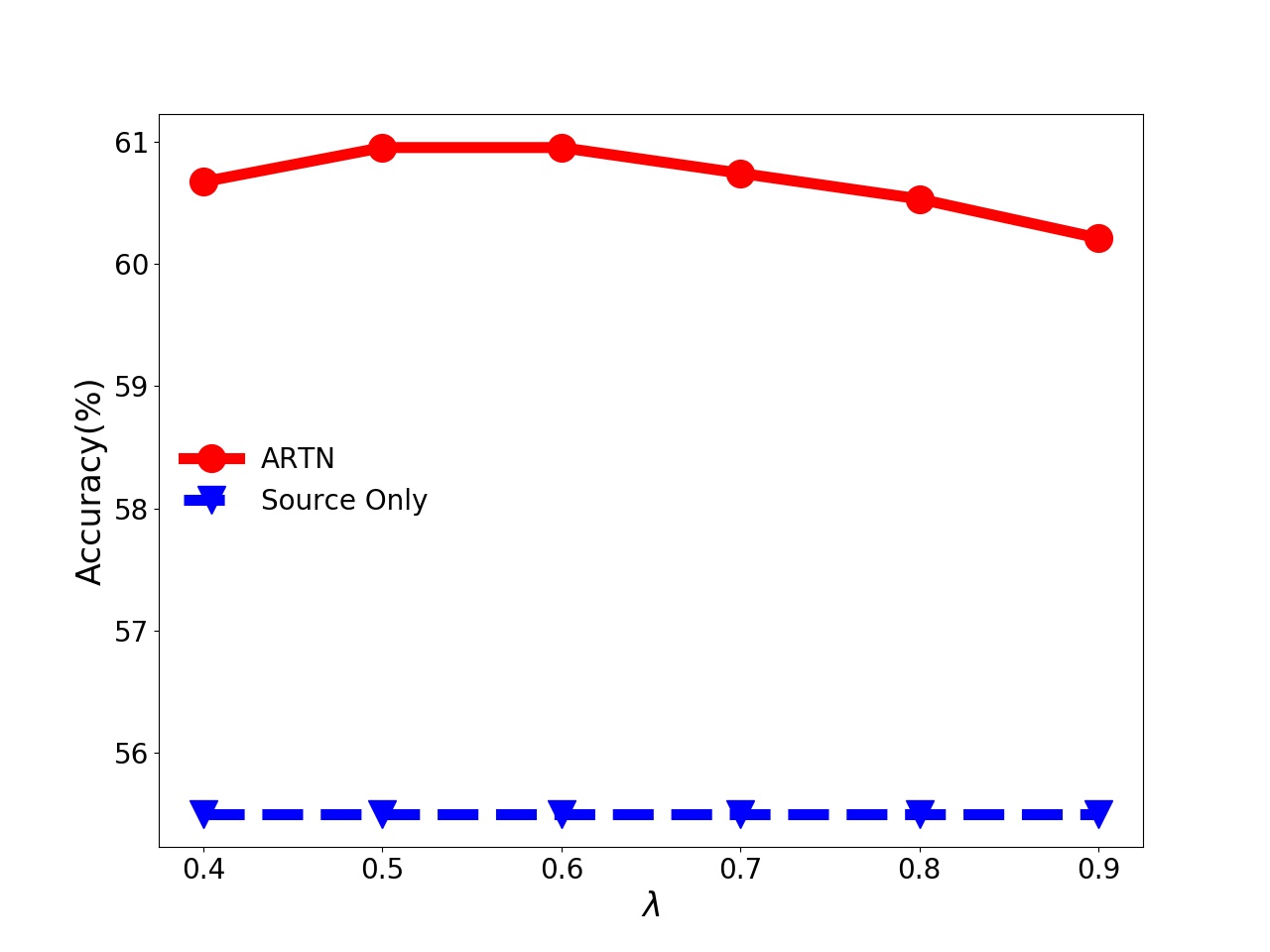}}
  \subfigure[D$\rightarrow$A]{
    \label{fig:subfig:DA} 
    \includegraphics[width=0.48\columnwidth]{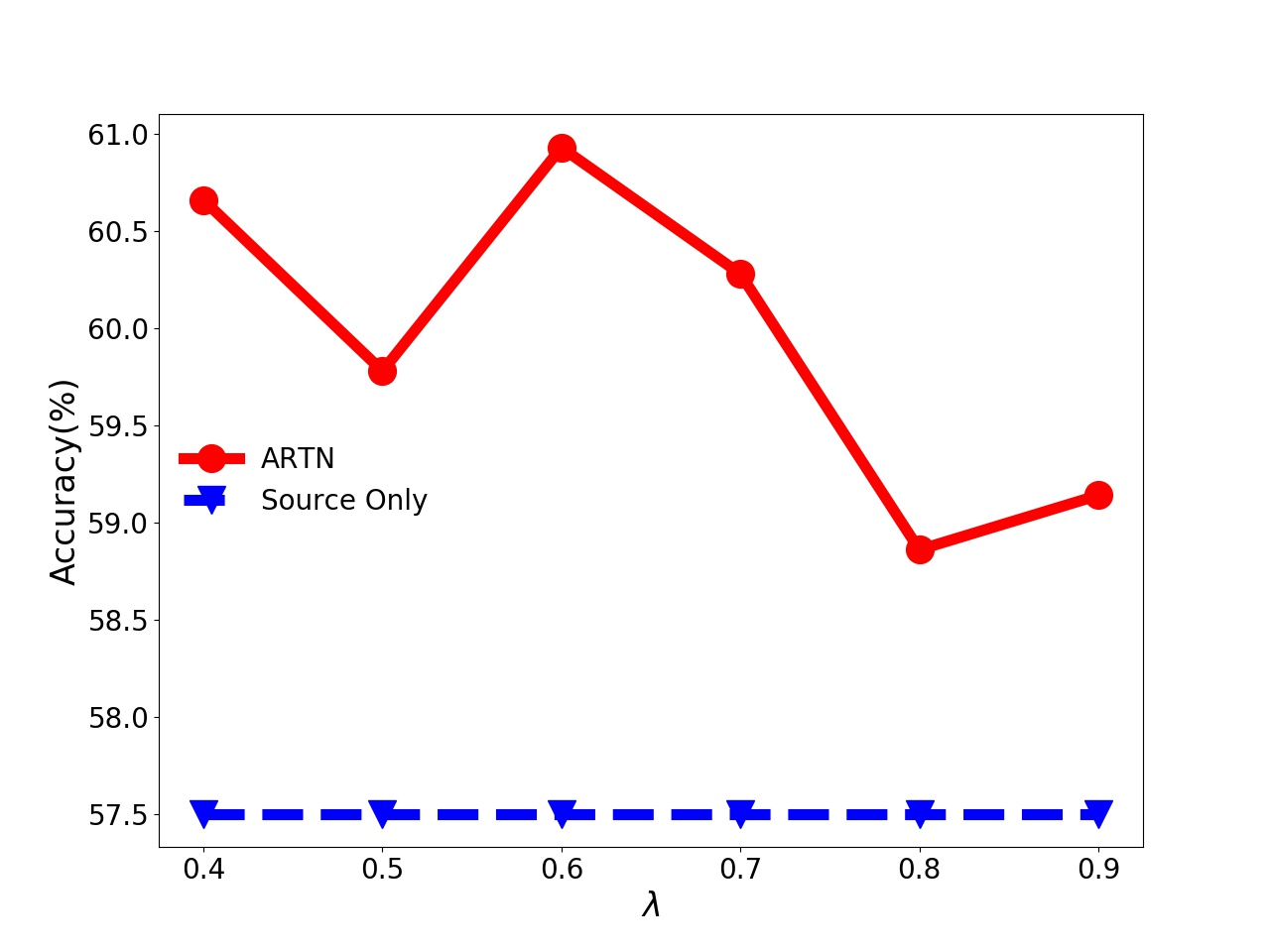}}  
\caption{Sensitivity of $\lambda$ in task A$\rightarrow$W, A$\rightarrow$D,W$\rightarrow$A, and D$\rightarrow$A. Dashed lines show results of adaptation-free method.}
  \label{fig:sens} 
\end{figure}

\subsection{Parameter Sensitivity}

In this experiment, we investigate how parameter $\lambda$ affects the performance of our model.  In order to make the results convincing, we test our model on tasks {\bf A}$\rightarrow${\bf W}, {\bf A}$\rightarrow${\bf D}, {\bf W}$\rightarrow${\bf A}, and {\bf D}$\rightarrow${\bf A} to acquire the variation of transfer classification performance as $\lambda\in\{0.4, 0.5, 0.6, 0.7, 0.8, 0.9\}$. Note that other settings are the same with those of the image classification experiment. In Fig.~\ref{fig:sens}, a detailed illustration is given.

The results in three of four tasks, {\bf A}$\rightarrow${\bf W}, {\bf A}$\rightarrow${\bf D} and {\bf W}$\rightarrow${\bf A} exhibit the same trend that the accuracy of ARTN is almost stable as $\lambda$ varies. Only in {\bf W}$\rightarrow${\bf A}, the accuracy fluctuates slightly with the variation of $\lambda$. Moreover, in the range of our settings, ARTN is always better than the model without adaptation and also better than most methods in Table~\ref{tab:table2}. This confirms the belief that ARTN is robust as $\lambda$ changes, which means the proposed method is no need for tuning hyper parameters subtly.

\section{Conclusion}
\label{section5}

We propose a novel unsupervised domain adaptation model based on adversarial learning. Different from previous adversarial adaptation models which rely on extracting domain-invariant representations, our model adds a feature-shared transform network to directly map features from a source domain to the space of target features. Furthermore, we add a regularization term to help strengthen its performance. Experimental results clearly demonstrate that the proposed model can match different domains effectively and is comparable with the state-of-the-art methods.


%

%


%
%

\ifCLASSOPTIONcaptionsoff
  \newpage
\fi



\bibliographystyle{IEEEtran}
\bibliography{IEEEabrv,tnnls}

\end{document}